\documentclass[journal]{IEEEtran} 
\usepackage{amsmath,amsfonts}
\usepackage{algorithmic}
\usepackage{algorithm}
\usepackage{array}
\usepackage[caption=false,font=normalsize,labelfont=sf,textfont=sf]{subfig}
\usepackage{textcomp}
\usepackage{stfloats}
\usepackage{url}
\usepackage{verbatim}
\usepackage{graphicx}
\usepackage{cite}
\usepackage{lipsum,adjustbox}
\usepackage{tikz}
\usetikzlibrary{trees, arrows.meta, positioning}
\usepackage{hyperref} %
\hypersetup{
    colorlinks,
    citecolor=black,
    filecolor=black,
    linkcolor=black,
    urlcolor=black,
    pdfauthor={},
    pdfsubject={},
    pdftitle={}
}
\usepackage[capitalize]{cleveref}

\usepackage{acro}
\usepackage{bm}
\usepackage{mathtools}
\usepackage{xcolor}
\usepackage{colortbl}     %
\usepackage{amssymb}
\usepackage{caption}
\usepackage[load-configurations = abbreviations]{siunitx}
\usepackage[inline]{enumitem}
\DeclareAcronym{ASL}{short = ASL, long = Autonomous Systems Lab}
\DeclareAcronym{OMAV}{short = robot, long = Omnidirectional Aerial Robot}
\DeclareAcronym{MAV}{short = MAV, long = Micro Aerial Vehicle}
\DeclareAcronym{DOF}{short = DOF, long = degrees of freedom}
\DeclareAcronym{PBC}{short = PBC, long = passivity-based control}
\DeclareAcronym{PH}{short = PH, long = Port-Hamiltonian}
\DeclareAcronym{NDT}{short = NDT, long = non-destructive testing}
\DeclareAcronym{PEMS}{short = PEMS, long = Power and Energy Monitoring System}
\DeclareAcronym{WTC}{short = WTC, long = wrench tracking controller}
\DeclareAcronym{MBE}{short = MBE, long = momentum-based wrench estimator}
\DeclareAcronym{ASIC}{short = ASIC, long = Axis-Selective Impedance Control}
\DeclareAcronym{MPC}{short = MPC, long = Model Predictive Control}
\DeclareAcronym{APhI}{short = APhI, long = Aerial Physical Interaction}
\DeclareAcronym{LLE}{short = LLE, long = Largest Lyapunov Exponent}
\DeclareAcronym{ICBF}{short = ICBF, long = Integral Control Barrier Functions}
\DeclareAcronym{CBF}{short = CBF, long = Control Barrier Functions}
\DeclareAcronym{COM}{short = CoM, long = Center of Mass}
\DeclareAcronym{AM}{short = AM, long = Aerial Manipulator}
\DeclareAcronym{IK}{short = IK, long = Inverse Kinematics}
\DeclareAcronym{IMU}{short = IMU, long = Inertial Measurement Unit}
\DeclareAcronym{FT}{short = F/T, long = force and torque, short-indefinite = an, long-indefinite = a}
\DeclareAcronym{LQR}{short = LQR, long = Linear Quadratic Regulator, short-indefinite = an, long-indefinite = a}
\DeclareAcronym{ESC}{short = ESC, long = Electronic Speed Controller, short-indefinite = an, long-indefinite = an}

\DeclareAcronym{apower}{short = PDA, long = Power Differential Allocation}
\DeclareAcronym{ageom}{short = GA, long = Geometric Allocation}
\DeclareAcronym{adiff}{short = ADA, long = Augmented Differential Allocation}
\DeclareAcronym{adiffold}{short = DA, long = Differential Allocation}
\DeclareAcronym{anorm}{short = NO, long = NO}
\DeclareAcronym{asecond}{short = NDA, long = Normalized Differential Allocation}
\DeclareAcronym{anosecond}{short = NDAnN, long = Normalized Differential Allocation with no Nullspace objectives}

\DeclareSIUnit{\rpm}{RPM}

\newcommand{\positiveGreen}{\cellcolor[HTML]{1ca33e}}
\newcommand{\negativeRed}{\cellcolor[HTML]{a31c1c}}
\renewcommand{\vec}[1]{\bm{#1}}		%
\renewcommand{\[}{\left[}		%
\renewcommand{\]}{\right]}		%
\newcommand{\mat}[1]{\bm{#1}}		%
\newcommand{\nR}[1]{\mathbb{R}^{#1}}		%
\newcommand{\upperRomannumeral}[1]{\uppercase\expandafter{\romannumeral#1}}	%

\newcommand{\transpose}{^\top}

\newcommand{\wrench}{\bm{w}}

\newcommand{\propSpeedLow}{\omega_l}
\newcommand{\propSpeedHigh}{\omega_h}
\newcommand{\propSpeed}{\omega}
\newcommand{\propSpeedVec}{\boldsymbol{\propSpeed}}
\newcommand{\tiltAngle}{\alpha}
\newcommand{\tiltAngleVec}{\boldsymbol{\tiltAngle}}
\newcommand{\propAcc}{\dot{\omega}}
\newcommand{\propAccVec}{\boldsymbol{\propAcc}}
\newcommand{\tiltSpeed}{\dot{\alpha}}
\newcommand{\tiltSpeedVec}{\boldsymbol{\tiltSpeed}}

\newcommand{\jointPosVec}{\vec{q}}
\newcommand{\jointSpeedVec}{\dot{\jointPosVec}}
\newcommand{\jointPos}{q}

\newcommand{\jointPosErrVec}{\jointPosVec_e}
\newcommand{\jointSpeed}{\dot{\jointPos}}
\newcommand{\jointSpeedDes}{\dot{\jointPosVec}^*}
\newcommand{\allocationMat}{\mat{A}}
\newcommand{\Dalloc}{\mat{D}}
\newcommand{\armsNum}{N}
\newcommand{\actuatorsVec}{\vec{u}}
\newcommand{\actuators}{u}
\newcommand{\jerk}{\dot{\wrench}}
\newcommand{\jerkMatrix}{\mat{J}}

\newcommand{\pseudo}{^\dagger}
\newcommand{\pseudoWeight}{^\ddagger}
\newcommand{\weightMat}{\mat{W}}
\newcommand{\identity}{\mat{I}}
\newcommand{\jerkGain}{k_j}
\newcommand{\saturationGain}{k_s}
\renewcommand{\min}[1]{\underline{#1}}
\renewcommand{\max}[1]{\overline{#1}}
\newcommand{\normal}[1]{{#1}_\text{n}}
\newcommand{\scalingMat}{\mat{N}}
\newcommand{\scalingBias}{\vec{b}}
\newcommand{\dynGain}{\mat{K}}

\newcommand{\propInertia}{J}
\newcommand{\propVoltage}{V}
\newcommand{\propCurrent}{I}
\newcommand{\propDrag}{d}
\newcommand{\powerEfficiency}{\eta}

\usepackage[normalem]{ulem}
\definecolor{kitblueex}{RGB}{52,115,186}
\definecolor{kitred}{RGB}{187,25,23}

\begin{document}

\title{Allocation for Omnidirectional Aerial Robots:\\
    Incorporating Power Dynamics
}

\author{Eugenio Cuniato,~\IEEEmembership{Member,~IEEE,}
Mike Allenspach,~\IEEEmembership{Member,~IEEE,}
Thomas Stastny,~\IEEEmembership{Member,~IEEE,}
Helen Oleynikova,~\IEEEmembership{Member,~IEEE,}
Roland Siegwart,~\IEEEmembership{Fellow,~IEEE,}
Michael Pantic,~\IEEEmembership{Member,~IEEE}
\thanks{All authors are with the Autonomous Systems Lab (ASL), ETH Zurich.}
\thanks{Corresponding author: {\tt \footnotesize \href{mailto:ecuniato@ethz.ch}{ecuniato@ethz.ch}.}}%
\thanks{The research leading to these results has been supported by the AERO-TRAIN project, European Union's Horizon 2020 research and innovation program under the Marie Skłodowska-Curie grant agreement No 953454, the ETH RobotX Research Program and the Armasuisse Research Grant No 3200000100. The authors are solely responsible for its content.}%
}

\markboth{IEEE TRANSACTIONS ON ROBOTICS}%
{Shell \MakeLowercase{\textit{et al.}}: A Sample Article Using IEEEtran.cls for IEEE Journals}

\maketitle

\begin{abstract}
    Tilt-rotor aerial robots are more dynamic and versatile than fixed-rotor platforms, since the thrust vector and body orientation are decoupled.
    However, the coordination of servos and propellers (the \textit{allocation} problem) is not trivial, especially accounting for overactuation and actuator dynamics.
    We incrementally build and present three novel allocation methods for tilt-rotor aerial robots, comparing them to state-of-the-art methods on a real system performing dynamic maneuvers.
    We extend the state-of-the-art geometric allocation into a differential allocation, which uses the platform's redundancy and does not suffer from singularities.
    We expand it by incorporating actuator dynamics and propeller \textit{power dynamics}.
    These allow us to model dynamic propeller acceleration limits, bringing two main advantages: balancing propeller speed without the need of nullspace goals and allowing the platform to selectively turn-off propellers during flight, opening the door to new manipulation possibilities.
    We also use actuator dynamics and limits to normalize the allocation problem, making it easier to tune and allowing it to track 70\% faster trajectories than a geometric allocation.
\end{abstract}


\section{Introduction}\label{sec:introduction}

Omnidirectional aerial robots have enjoyed increasing interest in the aerial robotics community~\cite{tognonTRO}.
The actuation capabilities of these systems allow omnidirectional thrust generation, resulting in fully-decoupled translational and rotational dynamics.
Recent works on aerial physical interaction demonstrate the superiority of omnidirectional platforms over traditional underactuated ones, ensuring precise motion and interaction force control with simultaneous disturbance rejection~\cite{bodie2021tro}.
\begin{figure}[t]
    \includegraphics[width=\columnwidth]{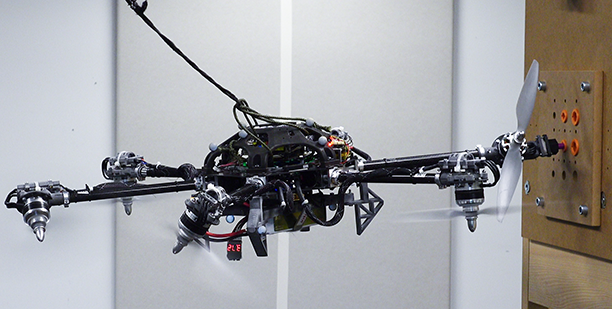}
    \caption{An omnidirectional tilt-rotor aerial robot screwing a bolt into a wall. The robot can smoothly transition its arms from flying mode (propellers on) to interaction (propellers off) with our novel allocation method. Video: \href{https://youtu.be/ybT80MyB9_Y}{youtu.be/ybT80MyB9\_Y}}
    \label{fig:omav}
\end{figure}
Depending on the specifics of the actuation, omnidirectional aerial robots can be classified as either \textit{fixed-rotor} or \textit{tilt-rotor} platforms.
Fixed-rotor systems feature propellers rigidly mounted at specific angles~\cite{park2018odar,tognon2019tilthex,ryll2019tilthex,brescianini2016omav}. While mechanically simple, they impose a fixed trade-off between efficiency and interaction capability, with no adaptability to mission or flight context.

Instead, tilt-rotor vehicles make use of dedicated actuators to modify the propeller orientation, depending on the task at hand~\cite{cuniato2023minivoliro,allenspach2020design,ryll2020fasthex,muellerpairtilt}.
This versatility strongly promotes their use in aerial interaction applications, where transitioning between an efficient navigation and a high-force interaction configuration might be required at different points throughout the mission~\cite{bodie2021tro,brunner2022mppi}.
However, it requires elaborate control allocation schemes to handle potential singularities, resolve actuator redundancy, and respect the difference in dynamic response between the tilt motors and the propellers.

These challenges have traditionally been addressed using complex, computationally expensive optimizers~\cite{ryll2018fasthex,ryll2020fasthex} or through heuristics in geometric allocation schemes~\cite{zhao2023versatile,nishio2023design}.
More recently, differential control allocation has emerged as a promising alternative~\cite{ryll2015fasthex,allenspach2020design}.
By formulating the allocation problem at the differential level, this approach incorporates not only geometric information but also the system’s kinematics.
However, despite its potential, significant gaps remain in the literature, as outlined in Section~\ref{sec:related_works}: (i) performance often depends on high-quality 6-DOF acceleration measurements; (ii) actuator dynamics and limits are usually neglected or oversimplified; (iii) propeller speed balancing for energy efficiency typically relies on complex nullspace objective switching; and (iv) thorough comparisons between differential and geometric allocation methods are still missing.

Aiming to address these limitations and enhance tilt-rotor robot capabilities, we propose and discuss different control allocation methodologies, experimentally compared on the platform shown in \cref{fig:omav}.
Our focus is on embedding actuator dynamics and limits directly into the allocation, leveraging overactuation without nullspace objectives, thereby enabling smooth in-flight propeller deactivation, opening new avenues for crash resilience and aerial manipulation.

\subsection{Contributions}\label{sec:contributions}
We incrementally address the challenges outlined above through the development of three novel actuator allocation schemes for tilt-rotor platforms:%

\begin{enumerate}
    \item We achieve up to 70\% faster trajectory tracking than geometric allocation by eliminating the need for acceleration feedback in differential allocation, enabling seamless integration with standard wrench-based control architectures.
    \item We make the allocation aware of its actuators limits and dynamics, balancing the utilization of our diverse actuators (propeller and servomotors) without requiring the manual tuning of a weighted pseudoinverse.
    \item We embed propeller power dynamics and model acceleration limits into the allocation description, providing a powerful new tool for optimizing flight efficiency and desired propeller speeds, without requiring elaborate nullspace objective switching.
    \item We demonstrate how these curves can be used to turn off propellers in flight, allowing the platform to control its propelling arms as manipulation tools, removing the need for additional manipulators.
\end{enumerate}

We begin the paper with a more in-depth review of related work in control allocation for tilt-rotor aerial robots in Section~\ref{sec:related_works}, further detailing the limitations outlined above.
The rest of the paper is organized as shown in \cref{fig:allocation_methods}, first introducing the overall system architecture in \cref{sec:system_architecture}, and then discussing and incrementally building up the five compared allocation methods (two state-of-the-art, three novel).
We then do a quantitative experimental comparison of all methods in \cref{sec:experimental_comparison} and show a qualitative demonstration of the ability of our proposed allocation to stop propellers in-flight for manipulation in \cref{sec:stopping_propellers}.

\begin{figure*}[t]
    \centering
    \begin{adjustbox}{width=\textwidth}
        \begin{tikzpicture}[
            edge from parent fork right,
            sibling distance=10mm,
            level distance=45mm,
            every node/.style={rectangle, draw, rounded corners, align=center, font=\normalfont, text width=35mm},
            edge from parent/.style={draw, -{Latex}, thick},
            grow=right,
            ]

            \node {\acl{ageom} (Sec.~\ref{sec:geometric_allocation})}
            child {node {\acl{adiffold} (Sec.~\ref{sec:differential_allocation})}
                    child {node {\acl{adiff} (Sec.~\ref{sec:augmented_allocation})}
                            child {node {\acl{asecond} (Sec.~\ref{sec:dynamics_allocation})}
                                    child {node {\acl{apower} (Sec.~\ref{sec:limit_allocation})}}
                                }
                        }
                };

        \end{tikzpicture}
    \end{adjustbox}
    \caption{The allocation methods introduced in this paper. The \acf{ageom} and \acf{adiffold} are the state-of-the-art geometric and differential allocation methods, respectively. The \acf{adiff} is a novel differential allocation method that does not require acceleration feedback. The \acf{asecond} normalizes the allocation problem with the actuator dynamics and limits, while the \acf{apower} introduces propeller power dynamics to minimize the use of nullspace commands and allow propeller deactivation in flight.}
    \label{fig:allocation_methods}
\end{figure*}

\section{Related Work}
\label{sec:related_works}
Considering their primary use-case as aerial manipulators, modern tilt-rotor aerial robots are mechanically designed to optimize omnidirectional force/torque envelopes, while exhibiting a dominant hover orientation for efficient flight.
As shown in \cite{allenspach2020design}, this optima is achieved by following a standard multicopter morphology, meaning all propellers are mounted in a co-planar and symmetric fashion on the main body.
Unlike standard underactuated multicopters, the direction of each propeller axis can be adjusted in flight, due to actuated tilting joints at the mounting point of the arms.
To achieve a desired force and torque acting on the body, we need to solve the \textit{allocation problem} of choosing the best distribution of propeller speeds and tilt arm directions across all the actuators.
In the following we summarize common solutions in the literature of tri-~\cite{cuniato2023minivoliro}, quad-~\cite{falconi2012,ryll2015fasthex,muellerpairtilt,qin2023design} and hexa-copter-like~\cite{rajappa2015fasthex,ryll2016fasthex,ryll2018fasthex,morbidi2018fasthex,ryll2020fasthex,bodie2021tro,bodie2020singularities,allenspach2020design} platforms.

\subsection{Hierarchical Allocation}
Motivated by the comparatively slow dynamics of the tilt arms, \cite{rajappa2015fasthex} proposes to optimize tilt angles along a given trajectory for efficient flight before the start of each mission, but then keeping them constant during execution.
To still ensure accurate tracking in the presence of disturbances, only the propeller speeds are updated online using a well-known matrix-inversion-based allocation scheme~\cite{johanssen2013allocation}.
Aside from requiring knowledge of the full task trajectory ahead of time, naively optimizing tilt angles for efficiency can result in singularities and loss of omnidirectionality~\cite{morbidi2018fasthex}.
Follow-up works~\cite{ryll2016fasthex,ryll2018fasthex,ryll2020fasthex} address this limitation by proposing to use an online tilt angle planner instead.
While also separately allocating tilt angles and rotor speeds, such a planner would aim to prevent singularities, without introducing excessive internal forces that reduce energy efficiency.
However, the choices of tilt angle in different flight situations seem to be very much task dependent and heuristically defined by the user.
In short, incorporating control authority and efficient flight while avoiding singularities is not trivial when tilt angle and propeller speed allocation are performed separately.

\subsection{Unified Geometric Allocation}
Alternatively, control allocation of both tilt angles and rotor speeds can be combined, resulting in a nonlinear and potentially high-dimensional system of equations to be solved.
Approaches presented in~\cite{falconi2012,bodie2020singularities,cuniato2023minivoliro,sugihara2024beatle,zhao2023versatile,nishio2023design} demonstrate how propeller force decomposition and trigonometric identities can produce a linear formulation suitable for matrix-inversion-based \textit{geometric allocation} schemes (see Sec~\ref{sec:geometric_allocation}).
Since the matrix becomes ill-conditioned when the system approaches a singular configuration, singularity cases must be analyzed and handled carefully.
Possible heuristics are proposed and experimentally validated in~\cite{bodie2020singularities}, for example biasing the desired arm angles to avoid numerical singularities.
However, the mathematical problem of mapping a desired wrench to tilt angles and rotor speeds still intrinsically suffers from singularities.

It should be noted that the cited works often neglect the \textit{different} dynamics between propellers and servomotors~\cite{bodie2021tro}, which can result in degraded disturbance rejection and dynamic tracking (see Sec.~\ref{sec:experimental_comparison}).

\subsection{Unified Differential Allocation}
To better coordinate the tilting motion and propeller dynamics, recent works propose to control the tilt angle \textit{velocities} and rotor \textit{accelerations} to recreate a desired change in total body force and torque, rather than forces and torques themselves.
In other words, to move the allocation problem to a higher differential level~\cite{ryll2015fasthex,allenspach2020design}.
This so-called \textit{differential allocation} (see Sec~\ref{sec:differential_allocation}) is not only more realistic in terms of actuator dynamics (as we no longer assume an instantaneous change of tilt angles and rotor speeds), but is also inherently robust to singularities~\cite{allenspach2020design}.
Furthermore, resolution of the actuator redundancy through nullspace exploitation is straightforward, for example to improve efficiency~\cite{ryll2015fasthex} or for consideration of mechanical constraints~\cite{allenspach2020design}.

While promising, existing differential allocation methods require measuring translational and rotational body accelerations.
Although acceleration feedback has been achieved on racing drones using onboard sensors~\cite{sun2022indi}, aerial manipulation platforms typically operate at much lower speeds and accelerations (\SI{20}{\meter\per\second}, 49\si{\meter\per\second^2} vs. \SI{1}{\meter\per\second}, \SI{2}{\meter\per\second^2}~\cite{zhao2023versatile,ryll2015fasthex,allenspach2020design}).
These gentler but precise maneuvers, combined with omnidirectional designs that decouple translational and rotational motion, result in minimal IMU excitation. Consequently, the signal-to-noise ratio is significantly reduced, making it difficult to obtain reliable acceleration measurements~\cite{sugihara2024beatle}.
Additionally, compared to racing drones, propeller speeds on aerial manipulation platforms tend to be more variable and operate closer to the system’s natural frequencies (e.g., \SI{22000}{\rpm} $\equiv$ \SI{366}{\hertz} vs. \SI{5000}{\rpm} $\equiv$ \SI{83}{\hertz}~\cite{ryll2020fasthex,allenspach2020design,ryll2015fasthex}), which complicates effective filtering.

Moreover, state-of-the-art differential allocation methods always require the use of posture nullspace objectives even just for free-flight, which keeps the nullspace always full and requires task switching in case a change in the nullspace goals is needed~\cite{notomista2020set} (e.g. from efficiency to mechanical constraints or manipulation tasks).

In this work, we iteratively build up three novel actuator allocation schemes for tilt-rotor platforms that integrate actuator dynamics, constraints, and potential secondary objectives into a unified differential allocation framework.
Unlike existing approaches, our methods do not rely on acceleration feedback and minimizes
the use of nullspace objective switching by directly encoding all relevant properties into the primary allocation problem.
We provide a quantitative experimental comparison against state-of-the-art geometric~\cite{bodie2020singularities} and differential~\cite{allenspach2020design} allocation strategies, demonstrating superior stability and dynamic tracking performance.

\section{System architecture}\label{sec:system_architecture}
\begin{figure}[t]
    \includegraphics[width=\columnwidth]{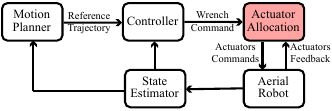}
    \caption{A general control architecture for tilt-rotor aerial robots. The pose controller compares the current odometry to a reference trajectory and produces a wrench command. The wrench command is finally allocated to the actuators of the robot, namely servomotors and propellers, to generate the desired wrench which drives the platform. In this work we focus on the allocation block, highlighted in red.}
    \label{fig:control_scheme}
\end{figure}
Tilt-rotor aerial robots have a number of propeller arms which can independently rotate around their axis using dedicated servomotors.
Each arm mounts a rotor at its tip, creating an independent thrust unit.
With three rotating propelling arms, the platform is fully-actuated, decoupling position and orientation control in space.
With more arms, it becomes overactuated, adding a nullspace that can be used to optimize further metrics apart from the main flight task.

For these robots, a common control architecture is shown in~\cref{fig:control_scheme}.
Given a set of sensors available on aerial robots (GPS, IMU, LiDAR, optical flow, ...), the state estimator provides an odometry measure~\cite{lanegger2023chasing}, which the controller~\cite{bodie2021tro} uses to generate a command wrench (forces and torques) to follow the current position and/or velocity reference.
Finally, the command wrench $\wrench \in \nR{6}$ is transformed into actuator commands for the platform's arm angles $\tiltAngleVec = \left[ \tiltAngle_0, ..., \tiltAngle_{\armsNum-1} \right]\transpose$ and propeller speeds $\propSpeedVec = \left[ \propSpeed_0, ..., \propSpeed_{\armsNum-1} \right]\transpose$ through the chosen allocation method, with $\armsNum \in \mathbb{Z}^{+}$ the number of arms.

In the following sections, we keep the motion planner, controller and state estimator fixed as in~\cite{bodie2021tro}, and focus on the actuator allocation problem.
A schematic of the allocation methods that will be introduced in the following sections can be seen in~\cref{fig:allocation_methods}.

\section{\acf{ageom}}\label{sec:geometric_allocation}
For the reader's convenience, here we reproduce the state-of-the-art geometric allocation method of~\cite{bodie2020singularities}, which only uses the platform's geometry to map the desired control wrench into propeller speeds and arm rotation angles.
\cite{bodie2020singularities} defines the allocation problem as
\begin{equation}\label{eq:geometric_allocation}
    \wrench = \allocationMat \actuatorsVec  \; , \; \actuatorsVec =
    \begin{bmatrix}
        \propSpeed_0\sin{\tiltAngle_0}                       \\
        \propSpeed_0\cos{\tiltAngle_0}                       \\
        ...                                                  \\
        \propSpeed_{\armsNum-1}\sin{\tiltAngle_{\armsNum-1}} \\
        \propSpeed_{\armsNum-1}\cos{\tiltAngle_{\armsNum-1}}
    \end{bmatrix},
\end{equation}
where $\allocationMat \in \nR{6\times 2\armsNum}$ represents a constant allocation matrix depending only on the system's geometry, while $\actuatorsVec \in \nR{2\armsNum}$ is the result of the allocation, obtained pseudoinverting the matrix $\allocationMat$.
Once the command vector $\actuatorsVec$ is generated, we extract the actual rotor speeds and tilt angles with
\begin{subequations}\label{eq:geometric_allocation_props_angles}
    \begin{align}
        \tiltAngle_i & = \operatorname{atan2}\left(\actuators_{2i}, \actuators_{2i+1} \right) \;,\; \forall i=0\,...\,\armsNum-1 \\
        \propSpeed_i & = \sqrt{\actuators_{2i}^2 + \actuators_{2i+1}^2} \;,\; \forall i=0\,...\,\armsNum-1.
    \end{align}
\end{subequations}
This method, which takes inspiration from the allocation of standard underactuated aerial robots, has the advantage of only requiring knowledge of the system's geometry.
However, it has a few limitations:
\begin{itemize}
    \item Since the allocation is purely geometric, it does not account for actuator dynamics or limits.
    \item The matrix $\allocationMat$ is constant and full rank, but the allocation problem can become singular depending on the desired wrench~\cite{bodie2020singularities}, i.e., it's not always possible to extract angles $\tiltAngleVec$ and rotor speeds $\propSpeedVec$ with~\cref{eq:geometric_allocation_props_angles}.
    \item The geometric allocation does not exploit the system's redundancy, only providing the solution with the minimum rotor speeds.
\end{itemize}
To overcome some of these limitations,~\cite{allenspach2020design} introduced a differential allocation method.

\section{\acf{adiffold}}\label{sec:differential_allocation}
The idea is to differentiate the relation in~\cref{eq:geometric_allocation} to map the time derivative of the wrench command (which we will from now on refer to as \textit{jerk}) to arm tilting speed and propeller acceleration, as
\begin{equation}\label{eq:jerk_allocation}
    \jerk = \allocationMat \Dalloc(\jointPosVec) \jointSpeedVec = \jerkMatrix(\jointPosVec) \jointSpeedVec,
\end{equation}
where $\jerk \in \nR{6}$ is the desired jerk to apply on the robot's body, $\jointPosVec = \left[ \tiltAngleVec\transpose, \propSpeedVec\transpose \right]\transpose\in\nR{2n}$ is the vector of joint states and $\Dalloc = \frac{\partial\actuatorsVec}{\partial\jointPosVec} \in \nR{2\armsNum\times2\armsNum}$ is the Jacobian of the geometric actuation vector $\actuatorsVec$.

The solution to this allocation problem is then
\begin{equation}\label{eq:jerk_allocation_solution}
    \jointSpeedVec = \jerkMatrix\pseudoWeight \jerk + (\identity_{2\armsNum} - \jerkMatrix\pseudoWeight\jerkMatrix)\jointSpeedDes,
\end{equation}
where the allocation's output is now a vector $\jointSpeedVec \in \nR{2\armsNum}$ of arm rotation speeds and rotor accelerations, $\jointSpeedDes \in \nR{2\armsNum}$ is a vector of desired joint velocities to exploit the system's overactuation by fulfilling secondary goals.

While distributing the desired jerk among the actuators, we must balance the usage of tilt angles and propellers, to avoid pushing either into saturation.
In~\cite{allenspach2020design}, the authors propose a weighted pseudoinverse
\begin{equation}\label{eq:weighted_pseudoinverse}
    \jerkMatrix\pseudoWeight = \weightMat^{-1} \jerkMatrix\transpose \left( \jerkMatrix \weightMat^{-1} \jerkMatrix\transpose \right)^{-1},
\end{equation}
with a positive-definite weight matrix $\weightMat \in \nR{2\armsNum\times2\armsNum}$ to distribute the desired jerk among the tilting arms and propellers.

This allocation method solves the singularity problem of the geometric solution in~\Cref{sec:geometric_allocation}~\cite{allenspach2020design}, while enabling use of the system's overactuation with $\jointSpeedDes$.
A possible choice for $\jointSpeedDes$ is to drive the propellers' speeds to a desired value (for example the hovering speed $\propSpeed^*$) with
\begin{equation}\label{eq:secondary_obj}
    \jointSpeedDes = -k_\propSpeed \[ \bm{0}_6 , (\propSpeed_0 - \propSpeed^*), ..., (\propSpeed_{\armsNum-1} - \propSpeed^*) \]^T,
\end{equation}
where $k_\propSpeed \in \nR{}$ is a positive gain.
Other choices, like actuator limits avoidance~\cite{ryll2015fasthex}, can be found in the literature of robot manipulators~\cite{siciliano2008springer}.

However, this method has two key limitations: it allocates a jerk command and relies on a heuristically chosen weighted pseudoinverse $\weightMat$ to balance arm and propeller use.
Both points are discussed in more detail in the following two subsections.
\subsection{Jerk vs. Wrench Control}
While the differential allocation requires a jerk command, wrench control is still the most common solution for aerial robots.
Indeed, it is the natural solution to the second-order rigid body dynamics, and it only requires position and velocity feedback, both commonly available on aerial robots.

The authors in~\cite{allenspach2020design} adopted a \ac{LQR} to produce the control jerk from acceleration feedback.
They designed two filters to obtain linear and angular acceleration feedback by filtering and differentiating \ac{IMU} data.

This solution permits the use of a differential allocation method, but its trajectory tracking performance shows no improvement over a standard PD controller with geometric allocation~\cite{allenspach2020design}.
We attribute this inefficiency to the \ac{LQR}’s strong model dependency and the limited quality of acceleration feedback.

As discussed in Section~\ref{sec:related_works}, low-excitation maneuvers reduce the IMU’s signal-to-noise ratio, complicating both filtering and differentiation of acceleration signals.\\
For this reason, in the following~\Cref{sec:augmented_allocation} we propose an augmented differential allocation, which does not require acceleration feedback and can be easily implemented with any standard wrench-based controller.

\subsection{Weighted Pseudoinverse} \label{sec:weighted_pseudo}
Differential allocation methods are common, for example, in the context of arm manipulators, where the behavior of the several actuators is similar and the joint speeds are usually in the same range.
However, in the case of tilt-rotor aerial robots, the behavior of the actuators is very different, as $\jointSpeedVec$ contains both propeller acceleration and rotation speed of the arms.
These two, apart from having different physical units and meanings, also have very different ranges, with the propeller accelerations usually around $\propAcc \sim 10^4 RPM/s$ and the arm speeds around $\tiltSpeed \sim1 rad/s$.

To keep the problem numerically stable, the differential allocation relies on a weighted pseudoinverse, which requires manual tuning of the weight matrix $\weightMat$ to balance the usage between tilt angles and propellers.
The matrix $\weightMat$ has a prominent effect on the aerial robot's behavior: too much weight on the tilt angles will push the allocation to use only the propellers, behaving the same way as a fixed multirotor, whereas too much weight on the propellers can lead the tilt angles' servomotors to rotate very fast and quickly saturate.
Finding a reliable $\weightMat$ often requires hours of in-flight tuning.

In~\Cref{sec:dynamics_allocation}, we propose a method to integrate actuator limits into the allocation problem to improve its numerical conditioning without the need of a hand-tuned weight matrix.
\begin{figure*}[t]
    \includegraphics[width=\textwidth]{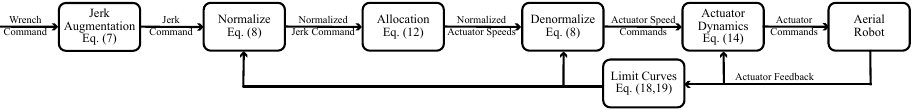}
    \caption{The proposed differential allocation scheme. We first augment the wrench command to generate a jerk command, we then normalize it and allocate into numerically stable normalized joint speeds. Finally, we invert the actuator dynamics to obtain the desired tilt angles and propeller speeds to command the Aerial Robot.}
    \label{fig:allocation_scheme}
\end{figure*}

\section{\acf{adiff}}\label{sec:augmented_allocation}
Here we propose Augmented Differential Allocation, to bring all the advantages of the Differential Allocation in~\cref{eq:jerk_allocation_solution} to any standard wrench-based controller, without the need for acceleration feedback.
We augment the input dynamics of our system in order to generate a jerk control command from a wrench command as
\begin{equation}\label{eq:jerk_augmentation}
    \jerk = \jerkGain \left(\wrench^d - \wrench(\jointPosVec)\right),
\end{equation}
where $\jerkGain \in \nR{}$ is a gain generating a jerk proportional to the error between the current wrench desired command $\wrench^d$ coming from a higher-level controller and the current wrench applied on the robot $\wrench(\jointPosVec)$ computed from the current actuators' state feedback.
This approach is much more reliable than IMU feedback and differentiation:
on one hand, it does not rely on platform- or sensor-dependent IMU filtering;
on the other, the actuators inherently act as low-pass filters, naturally providing a less noisy estimate.

Moreover, this jerk augmentation was already validated in the context of \ac{CBF} for ground~\cite{Ames2021} and aerial robots~\cite{cuniato2022power}.
However, it still requires tuning of the weight matrix $\weightMat$ in~\cref{eq:weighted_pseudoinverse} to balance the usage between tilt angles and propellers, which ultimately still relies on user heuristics obtained in flight tests.

\section{\acf{asecond}}\label{sec:dynamics_allocation}
To avoid heuristically tuning the weight matrix $\weightMat$, we propose to balance the actuators' usage by normalizing the allocation problem with the knowledge of the actuators' control limits, instead of using a weighted pseudoinverse.

\subsection{Actuator Control Limits}
To normalize the allocation problem, consider the minimum and maximum actuator rate vectors $\min{\jointSpeedVec} = \left[ \min{\tiltSpeedVec}\transpose, \min{\propAccVec}\transpose \right]\transpose$ and $\max{\jointSpeedVec} = \left[ \max{\tiltSpeedVec}\transpose, \max{\propAccVec}\transpose \right]\transpose$.
We refer with \textit{actuator rate} $\jointSpeedVec$ to angular velocities for the tilt angles and rotor accelerations for the propellers.
This is because the state of our actuators $\jointPosVec$ is composed of both tilt angles and rotor speeds, which can be regarded as an \textit{actuator position}.

We now normalize the actuator velocities with respect to these limits, mapping $\jointSpeed_i \in [\min{\jointSpeed}_i, \max{\jointSpeed}_i]$ to $\normal{\jointSpeed}{}_{,i} \in [-1,1]$ with
\begin{equation}\label{eq:normalized_velocity_vec}
    \normal{\jointSpeedVec} = \scalingMat \jointSpeedVec - \scalingBias,
\end{equation}
with a diagonal scaling matrix $\scalingMat \in \nR{2\armsNum\times2\armsNum}$ and a bias vector $\scalingBias \in \nR{2\armsNum}$, where
\begin{equation}\label{eq:N_b}
    N_{i,i} = \frac{2}{\max{\jointSpeed}_i - \min{\jointSpeed}_i}\;,\;
    b_i     =  \frac{\max{\jointSpeed}_i + \min{\jointSpeed}_i}{\max{\jointSpeed}_i - \min{\jointSpeed}_i} \;,\; \forall i=0,\dots,2\armsNum-1.
\end{equation}
We now rewrite the allocation problem in~\cref{eq:jerk_allocation} as a function of the normalized joint speeds $\normal{\jointSpeedVec}$ instead of the speeds themselves $\jointSpeedVec$.
Substituting~\cref{eq:normalized_velocity_vec} in the original differential allocation~\cref{eq:jerk_allocation} we obtain the normalized allocation
\begin{equation}\label{eq:normalized_jerk_allocation_extended}
    \jerk = \jerkMatrix(\jointPosVec) \scalingMat^{-1} \normal{\jointSpeedVec} + \jerkMatrix(\jointPosVec) \scalingMat^{-1} \scalingBias,
\end{equation}
which is equivalent to
\begin{equation}\label{eq:normalized_jerk_allocation}
    \normal{\jerk} = \normal{\jerkMatrix} \normal{\jointSpeedVec},
\end{equation}
where the normalized jerk command to allocate is now $\normal{\jerk} = \jerk - \jerkMatrix \scalingMat^{-1} \scalingBias \in \nR{6}$ and the normalized allocation matrix is $\normal{\jerkMatrix} = \jerkMatrix \scalingMat^{-1} \in \nR{2\armsNum\times2\armsNum}$.
Similarly to the original differential allocation, the solution to this problem is
\begin{subequations}\label{eq:dyn_jerk_allocation_solution}
    \begin{align}
        \normal{\jointSpeedVec}     & = \normal{\jerkMatrix}\pseudo \normal{\jerk} + (\identity_{2\armsNum} - \normal{\jerkMatrix}\pseudo\normal{\jerkMatrix})\normal{\jointSpeedDes}, \\
        \normal{\jerkMatrix}\pseudo & = \normal{\jerkMatrix}\transpose \left( \normal{\jerkMatrix}^{} \normal{\jerkMatrix}\transpose \right)^{-1}.
    \end{align}
\end{subequations}
Despite the similarity, this formulation has the advantage of not requiring manual tuning of a weighted pseudoinverse.
Instead, actuator balancing is naturally induced by their rate limits, as used in~\cref{eq:N_b}, which we identified experimentally on a test bench.

\subsection{Actuator Saturation}
\label{sec:actuator_saturation}

Despite normalizing the actuator commands between $[-1,1]$, given a normalized jerk $\normal{\jerk}$, the allocation might still output normalized joint velocity commands $\normal{\jointSpeedVec}$ outside of these boundaries.
This can happen if the required jerk command is too high, for example when requiring very rapid maneuvers.
We saturate the joint velocities between $[-1,1]$, by linearly scaling down the velocity vector with a quantity $\saturationGain < 1$ as
\begin{equation}
    \text{sat}(\normal{\jointSpeedVec}) = \saturationGain \normal{\jointSpeedVec}.
\end{equation}
The desired gain is $\saturationGain = \frac{1}{|\normal{\jointSpeed}{}_{,b}|}$, where $\normal{\jointSpeed}{}_{,b}$ is the biggest element of the joint speeds vector.
Notice that since all the actuator commands in $\normal{\jointSpeedVec}$ are scaled by the same quantity $\saturationGain$ and the relation in~\cref{eq:normalized_jerk_allocation} is linear, this corresponds to generating a jerk $\text{sat}(\normal{\jerk})$ which has the same direction of the original $\normal{\jerk}$ but scaled in magnitude.
While such scaling is well established in robotic manipulator control~\cite{siciliano2008springer,fabrizio2015nullspace}, it remains unique in aerial robotics. Even in the few works that adopt jerk-level allocation, actuator saturation is handled through simple clipping~\cite{allenspach2020design} or heuristic null-space strategies without feasibility guarantees~\cite{ryll2015fasthex}.
In contrast, our method enables a more intuitive and systematic approach by leveraging proven techniques from robotic manipulators, highlighting the conceptual similarity between omnidirectional aerial robots and fully-actuated robot arms.

\subsection{Actuator Dynamics}\label{sec:actuator_dynamics}
The described allocation procedure yields actuator rate commands $\jointSpeedVec$ that aim to realize the desired jerk $\jerk$.
While the standard approach is to integrate these rates to obtain actuator setpoints $\jointPosVec$~\cite{allenspach2020design,ryll2015fasthex}, we leverage the differential formulation to incorporate not only actuator velocity \textit{limits} but also their \textit{dynamics}.

Using data from real flights, we identify a first-order actuator model via least-squares fitting:
\begin{equation}\label{eq:first_order_dynamics}
    \jointSpeedVec = -\dynGain \jointPosErrVec,
\end{equation}
where $\dynGain \in \nR{2\armsNum\times2\armsNum}$ is a positive-definite diagonal gain matrix, and $\jointPosErrVec \in \nR{2\armsNum}$ represents angular and rotational speed errors for tilt arms and propellers, respectively.

This first-order model aligns naturally with our rate-based allocation. In contrast, second-order dynamics would require differentiating~\cref{eq:dyn_jerk_allocation_solution} and access to propeller acceleration feedback, typically unavailable from current \acp{ESC}.

We thus compute desired actuator rates by solving the allocation as before and then invert~\cref{eq:first_order_dynamics} to obtain dynamically feasible actuator setpoints.
All the steps of the proposed allocation process are summarized in~\cref{fig:allocation_scheme}.

\section{\acf{apower}}\label{sec:limit_allocation}
Differential allocation controls the actuators' rate (propeller acceleration and tilt arm speed), but not directly their states (propeller speed and tilt arm angles).
This might not be a problem for the tilt arms, which could virtually rotate indefinitely depending on their mechanical implementation.

However, propellers' speed are limited by the maximum power that the \acp{ESC} can provide, propeller's inertia and drag coefficient.
Aside from these constraints, balanced thrust distribution is particularly important for tiltrotor aerial vehicles.
At high pitch or roll angles, some propellers may spin significantly faster than others. If not \textit{actively} rebalanced upon returning to level hover, the system can remain in a power-inefficient state, unnecessarily pushing actuators toward saturation.

The differential allocations introduced so far rely on nullspace goals such as~\cref{eq:secondary_obj} to keep the speed balanced and avoiding propellers to hit the limit.
Here we introduce the propeller power dynamics:
a model in which acceleration limits in Eq.~\eqref{eq:normalized_velocity_vec} adapt dynamically based on current propeller speeds.
This allows for both enforcing speed constraints and encouraging speed equalization directly within the allocation, without requiring an explicit nullspace objective.
This contrasts with established approaches in robotic manipulation~\cite{siciliano2008springer,fabrizio2015nullspace}, which commonly introduce a secondary nullspace allocation goal to optimize the robot's posture.
Rather than aiming to outperform such nullspace strategies, our method preserves the nullspace for aerial-robotics-specific objectives such as internal force regulation~\cite{siciliano2008springer}, mechanical considerations~\cite{allenspach2020design}, or manipulation-oriented tasks (see \cref{sec:stopping_propellers}).

We derive the power dynamics in two steps:
First, in \cref{sec:real_propeller_curves}, we identify the physical limits imposed by the ESCs, propeller drag, and inertia.
Then, in \cref{sec:proposed_power_limits}, we propose a limit curve formulation that respects these mechanical constraints while also encouraging speed equalization.

\subsection{Real Propeller Curves}
\label{sec:real_propeller_curves}
\begin{figure}[t]
    \includegraphics[width=\columnwidth]{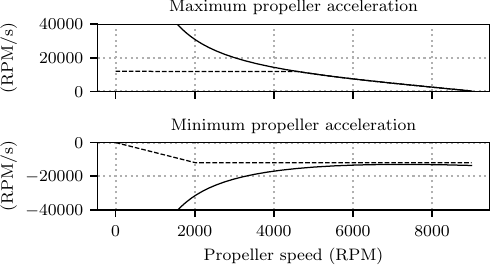}
    \caption{Example acceleration limit curves (solid lines) from the propellers power balance in~\cref{eq:power_equilibrium}. The dashed lines represent the saturated versions of the curves, where the maximum and minimum accelerations have been capped to represent the internal software ESC limits. For very low propeller speeds, the drag torque has a very low effect, leading to very high acceleration and deceleration limits. As speed increases, the quadratic drag force overtakes the electrical torque, leading to a reduction in the maximum acceleration and deceleration, until the acceleration becomes zero when the propeller reaches the maximum speed. The curves were obtained with $\powerEfficiency=0.8$, $\propVoltage = 23V$, $\max{\propCurrent} = -\min{\propCurrent} = 17A$, $\propInertia = 4.5e^{-4} Kg m^2$, $\propDrag = 3.5e^{-7} N m s^2$ and $\max{\propAcc}=-\min{\propAcc}=1.3e^4 RPM/s$.}
    \label{fig:propeller_example_limit_curves}
\end{figure}
The propeller acceleration curves for a generic propeller motor can be obtained through the power equilibrium equation.
Specifically, the input electrical power must be equal to the power dissipated to fight the propeller's inertia and drag torque.
This translates to
\begin{equation}\label{eq:power_equilibrium}
    \powerEfficiency\propVoltage\propCurrent = \propInertia\propSpeed\propAcc + \propDrag\propSpeed^3,
\end{equation}
where all the quantities are scalar and $\powerEfficiency < 1$ is the electrical efficiency of the motor, $\propVoltage$ is the voltage provided to the \ac{ESC}, $\propCurrent$ is the current absorbed by the motor, $\propInertia$ is the propeller and rotor combined inertia and $\propDrag$ is the drag coefficient.
If we consider $\max{\propCurrent}$ the maximum current the \ac{ESC} is able to provide, we can rewrite~\cref{eq:power_equilibrium} to obtain the maximum propeller acceleration
\begin{equation}\label{eq:propeller_acceleration_max}
    \max{\propAcc} = \frac{\powerEfficiency\propVoltage\max{\propCurrent}}{\propInertia} \frac{1}{\propSpeed} - \frac{\propDrag}{\propInertia}\propSpeed^2.
\end{equation}
Similarly, the minimum acceleration $\min{\propAcc}$ can be obtained by substituting the minimum current $\min{\propCurrent}$ in~\cref{eq:propeller_acceleration_max}.

An example of these curves for a typical propeller-\ac{ESC} combination is shown in~\cref{fig:propeller_example_limit_curves} (solid line).
While the curves fit the real propeller behavior well at high speeds where the actual physical power expenditure is dominating, at lower speeds the \ac{ESC} control algorithm and other losses are more apparent:
\begin{itemize}
    \item Most multirotors are not designed to have their propellers spin backward, which means that the minimum acceleration $\min{\propAcc}$ should actually go to zero as the speed approaches the minimum $\min{\propSpeed}$ (usually zero).
    \item As shown in~\cref{fig:propeller_example_limit_curves}, the maximum acceleration would ideally approach infinity as the propeller speed approaches zero. In practice, however, it is bounded by the internal controller and hardware design of the \ac{ESC}. As a result, the maximum acceleration should also saturate to $\max{\propAcc}$ as the speed approaches the minimum $\min{\propSpeed}$.
\end{itemize}
We illustrate these modified acceleration limits in~\cref{fig:propeller_example_limit_curves} as dashed lines.

\subsection{Propellers' Equilibrium Speed and Proposed Curves}\label{sec:proposed_power_limits}
\begin{figure}[t]
    \includegraphics[width=\columnwidth]{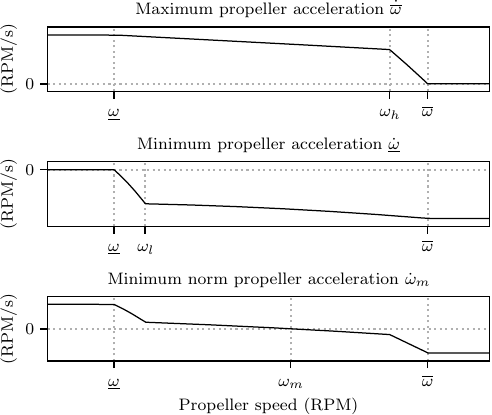}
    \caption{Proposed propeller limit curves. This approximation allows arbitrarily choosing the equilibrium speed $\propSpeed_m$ of the propellers. The first two rows are the maximum $\max{\propAcc}$ and minimum $\min{\propAcc}$ acceleration curves as a function of the rotor speed. The last row is the mean $\propAcc_m = \frac{\max{\propAcc}+\min{\propAcc}}{2}$.}
    \label{fig:proposed_limit_curves}
\end{figure}
We aim to design propeller acceleration limit curves that not only respect the mechanical constrains in \cref{fig:propeller_example_limit_curves} but also promote speed equalization among propellers.
This is achieved by leveraging the fact that the minimum norm solution of the allocation problem in \cref{eq:dyn_jerk_allocation_solution} corresponds to the midpoint between the acceleration limits, i.e. $\normal{\jointSpeedVec} \rightarrow \bm{0}\Rightarrow\propAcc \rightarrow \frac{\max{\propAcc}+\min{\propAcc}}{2}$.

We define the associated speed as the equilibrium speed $\propSpeed_m$

\begin{equation} \label{eq:equilibrium_speed}
    \propAcc_m(\propSpeed_m) = \frac{\max{\propAcc}(\propSpeed_m)+\min{\propAcc}(\propSpeed_m)}{2} = 0 .
\end{equation}
The key idea is to use this equilibrium speed as a tuning parameter when shaping the maximum and minimum acceleration curves.
Since the minimum norm solution $\propAcc_m$ is positive for speeds below $\propSpeed_m$ and negative for speeds above $\propSpeed_m$, the allocation passively drives each propeller toward $\propSpeed_m$, removing the need for secondary nullspace goals to balance propeller speeds.

By tuning $\omega_m$, we can prioritize different behaviors: setting it low favors energy efficiency, while choosing a value above nominal hovering speed encourages greater internal force generation and thus faster dynamic response.
While we only optimize for energy efficiency in our work, as an overactuated robot, the posture (identified by tilt angles and rotor speeds) affects the manipulability (or \textit{dexterity index}~\cite{siciliano2008springer}), which is the ability of effectively generating the desired jerk command.

Keeping these requirements in mind, we propose to model the maximum and minimum acceleration curves as composed of two pieces each, as shown in~\cref{fig:proposed_limit_curves}
\begin{equation}\label{eq:modified_acceleration_max}
    \max{\propAcc}(\propSpeed) =
    \begin{cases}
        c_{0,0}\propSpeed + c_{0,1}\propSpeed^2 + c_{0,2}, & \forall \propSpeed \in [\min{\propSpeed}, \propSpeedHigh] \\
        c_{1,0}\propSpeed^2 + c_{1,1},                     & \forall \propSpeed \in [\propSpeedHigh, \max{\propSpeed}]
    \end{cases}
\end{equation}
\begin{equation}\label{eq:modified_acceleration_min}
    \min{\propAcc}(\propSpeed) =
    \begin{cases}
        c_{2,0}\propSpeed^2 + c_{2,1}, \phantom{+ c_{2,2}} & \forall \propSpeed \in [\min{\propSpeed}, \propSpeedLow] \\
        c_{3,0}\propSpeed^2 + c_{3,1},                     & \forall \propSpeed \in [\propSpeedLow, \max{\propSpeed}]
    \end{cases}
\end{equation}
where $\propSpeedHigh$ and $\propSpeedLow$ are two arbitrary propeller speeds in which we start ramping down the maximum acceleration and ramping up the minimum acceleration, respectively.

With their $9$ coefficients, the curves are uniquely identified once the maximum/minimum speeds and accelerations are chosen, together with the equilibrium speed.
Recomputing the new limits requires the solution of a small linear system of equations which is detailed in~\Cref{sec:power_curves_math}. Then, the new limits can be applied by just recomputing the scaling matrix and bias vector in~\cref{eq:N_b}.
\begin{figure}[b]
    \includegraphics[width=\columnwidth]{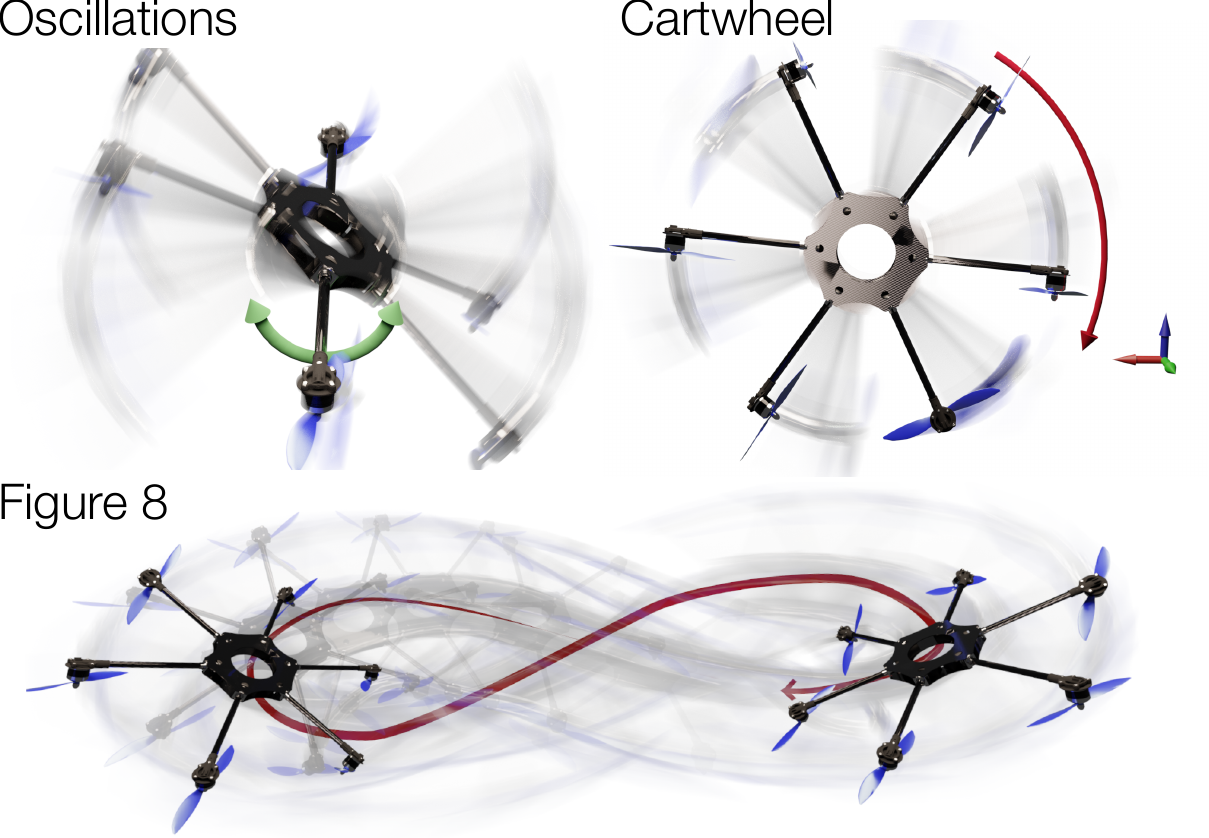}
    \caption{Illustration of evaluation trajectories. Top Left: Sinusoidal motion around body axes with increasingly faster cycle times, the roll motion is visualized. Top Right: Cartwheel trajectory, in which the robot is commanded to fly upright in the vertical plane and rotate slowly around the perpendicular axis. Bottom: Figure 8 that smoothly excites all axes -- similar to the well-known ``Lazy 8" in aviation.}
    \label{fig:vel_time_traj}
\end{figure}
\section{Experimental Evaluation}\label{sec:experimental_comparison}
\begin{figure}[t]
    \includegraphics[width=\columnwidth]{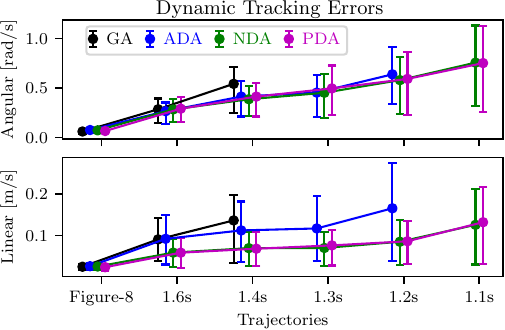}
    \caption{Mean and quartiles of the linear and angular tracking velocity errors achieved by the geometric (\ac{ageom}) and differential (\ac{adiff}, \ac{asecond}, \ac{apower}) allocations. When data from an allocation is missing for a specific trajectory, it means that the method failed to complete the trajectory. The use of actuator dynamics allows \acs{apower} and \acs{asecond} to complete all trajectories.}
    \label{fig:trajectory_tracking_evolution}
\end{figure}
Here we evaluate the proposed differential allocation methods (\ac{adiff}, \ac{asecond}, \ac{apower}) against the geometric method \ac{ageom}.
The following subsections showcase the effects of the respective enumerated contributions in~\ref{sec:contributions}:
\begin{enumerate}
    \item Section~\ref{sec:trajectory_tracking} compares tracking performance on both slow and fast oscillating trajectories between the geometric allocation (\ac{ageom}), differential allocations that use a weighted pseudoinverse to balance actuator commands (\ac{adiff}) and the ones using actuator limit normalization (\ac{asecond}, \ac{apower}).
    \item Section~\ref{sec:power_consumption} highlights the importance of actuator saturation handling while using differential allocation methods and its influence on power consumption, particularly comparing the use of nullspace commands \ac{asecond} and propeller curves \ac{apower}, vs. the total absence of any propeller speed regulation techniques \acs{anosecond}.
    \item Section~\ref{sec:cartwheel} evaluates behaviors while flying through allocation singularities, showing the robustness of differential over geometric allocations.
    \item Section~\ref{sec:stopping_propellers} demonstrates how propeller curves (\ac{apower}) can smoothly turn off propellers in flight and use the corresponding arms as proper manipulators.
\end{enumerate}
\subsection{Trajectory tracking}\label{sec:trajectory_tracking}
We fly the robot through seven different trajectories: a slow Figure-8 and six increasingly fast and dynamic trajectories that involve different rotations on the spot (\cref{fig:vel_time_traj}).
In~\cref{tab:traj_success} we summarize the success of each allocation method in completing the desired trajectories, eventually bringing all of them to failure.
A run is considered successful as long as the controller does not diverge, i.e. the errors are bounded.
We also show their dynamic tracking errors (linear and angular) in~\cref{fig:trajectory_tracking_evolution}.
Here we focus specifically on dynamic tracking errors as we are interested in the dynamic performance of these allocation methods, but a full statistical overview of the tracking errors (including position and attitude) is available in~\cref{fig:boxplots_errors} later in the Appendix.

The first to fail is \ac{ageom}, which completes the Figure-8 and only the two slowest oscillating trajectories.
This was expected as this method assumes instantaneous control of the actuators, without any knowledge of actuator dynamics and limits.

The second method to fail is \ac{adiff}, which completes most of the proposed trajectories.
Despite not having knowledge of the exact actuator dynamics and limits, this method still performs comparably well, although it is highly dependent on a careful tuning of the weight matrix $\weightMat$ to balance the usage of the actuators.

Once the allocation problem is normalized—by incorporating actuator dynamics and limits and removing the weight matrix $\weightMat$—both \ac{asecond} (with nullspace objectives) and \ac{apower} (using propeller power curves) demonstrate the highest tracking performance.
Notably, \ac{apower} achieves this while consistently keeping the nullspace available during nominal flight.
This leads to two key benefits:
first, it eliminates the need for task-switching or prioritization strategies when transitioning from free flight to manipulation tasks (\Cref{sec:stopping_propellers}), since additional objectives naturally exploit the already free nullspace;
second, it avoids the need for in-flight tuning of the nullspace goal, as the power curves are inherently derived from the physical characteristics of the rotors.

To assess whether the tracking errors between the two methods differ significantly, we perform Welch’s unequal variances t-test.
We set the null hypothesis that the velocity error distributions from~\cref{fig:trajectory_tracking_evolution} have equal mean values, rejecting it if the p-value is below $0.05$.
The results indicate no statistically significant difference between the two approaches, with p-values of $0.106$ for linear and $0.788$ for angular tracking errors.

In summary, \ac{apower} offers robust tracking performance comparable to \ac{asecond}, while providing practical advantages through nullspace availability and physically grounded design.

\begin{table}[t]
    \renewcommand*{\arraystretch}{1.0}
    \centering
    \begin{tabular}{|c || c | c | c | c | c | c | c |}
        \hline
                      & Fig-8          & \qty{1.6}{\second} & \qty{1.4}{\second} & \qty{1.3}{\second} & \qty{1.2}{\second} & \qty{1.1}{\second} & \qty{1.0}{\second} \\ [0.5ex]
        \hline\hline

        \acs{ageom}   & \positiveGreen & \positiveGreen     & \positiveGreen     & \negativeRed       & \negativeRed       & \negativeRed       & \negativeRed       \\ [0.5ex]
        \hline
        \acs{adiff}   & \positiveGreen & \positiveGreen     & \positiveGreen     & \positiveGreen     & \positiveGreen     & \negativeRed       & \negativeRed       \\ [0.5ex]
        \hline
        \acs{asecond} & \positiveGreen & \positiveGreen     & \positiveGreen     & \positiveGreen     & \positiveGreen     & \positiveGreen     & \negativeRed       \\ [0.5ex]
        \hline
        \acs{apower}  & \positiveGreen & \positiveGreen     & \positiveGreen     & \positiveGreen     & \positiveGreen     & \positiveGreen     & \negativeRed       \\
        \hline\hline
        Ang. vel.     & 0.5            & 2.3                & 2.8                & 3.2                & 3.5                & 4.0                & -                  \\ [0.5ex]
        \hline
        Lin. vel.     & 0.4            & 0.17               & 0.20               & 0.25               & 0.35               & 0.6                & -                  \\ [0.5ex]
        \hline
    \end{tabular}
    \caption{The table summarizes whether an allocation method was successful in completing the desired trajectory (green) or not (red). Allocation methods are on the rows while the trajectories are on the columns slowest (left) to fastest (right). The best methods (\acs{apower} and \acs{asecond}) require the use of actuator dynamics and normalized actuator limits. The last two rows show the maximum linear (\qty{}{\meter\per\second}) and angular (\qty{}{\radian\per\second}) body velocity norms reached during the corresponding trajectory.}\label{tab:traj_success}
\end{table}
\begin{figure}[t]
    \includegraphics[width=\columnwidth]{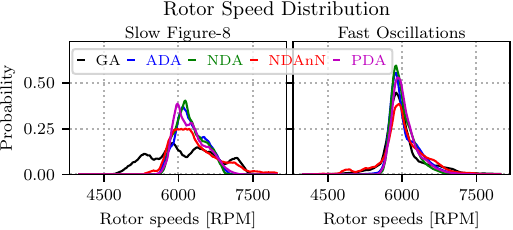}
    \caption{Comparison of rotor speeds distributions between the different allocation methods. When the secondary objective is not present (\ac{anosecond}) or in case of the geometric allocation (\ac{ageom}), the rotor speeds distribution is wider and the propellers are pushed closer to their limits.}
    \label{fig:rotor_speeds_comparison}
\end{figure}

\subsection{Balancing propellers' speed and power consumption}\label{sec:power_consumption}

During flight, the speed of the propellers needs to be carefully regulated to avoid propeller saturation and unnecessarily high power consumption.
While the geometric allocation (\ac{ageom}) by definition always minimizes the propeller speed to obtain the desired control wrench, differential methods (\ac{adiff}, \ac{asecond}, \ac{apower}) only operate in terms of propeller accelerations to generate the desired jerk.
In our experiments, we try to keep the propellers close to the $5800 RPM$ hovering speed of our system, either through nullspace objectives (\ac{adiff}, \ac{asecond}) or the proposed propeller curves (\ac{apower}).

If neither of these are present, the speeds can slowly drift towards the propeller upper limit of $8800 RPM$.
To demonstrate this, in~\cref{fig:rotor_speeds_comparison} we compare the rotor speeds distributions between the different allocation methods.
We additionally evaluate one of the best methods, \ac{asecond}, but without the nullspace secondary objective that regulates propeller speeds, obtaining \ac{anosecond}.
We can see \ac{anosecond}'s speed distribution having an upper tail towards higher rotor speeds during the slow Figure-8 trajectory, while the other differential methods manage to keep the speeds close to $5800 RPM$.
\ac{ageom} also has a wide speed distribution as it regulates in terms of \emph{energy} efficiency, without regard for actuator limits.

\begin{figure}[b]
    \includegraphics[width=\columnwidth]{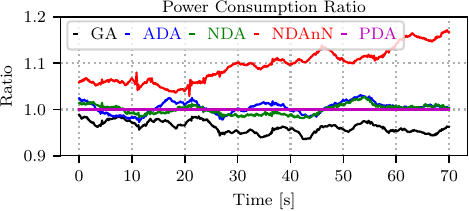}
    \caption{Power usage ratio between \ac{apower} and the other allocation methods during the slow Figure-8 trajectory. \ac{ageom} is up to 6\% more efficient than the other differential methods, which have similar consumptions. \acs{anosecond} increases the average propeller speeds over time, leading to a power consumption up to 18\% higher than the \acs{apower} method. The power consumption is computed using \cref{eq:power_equilibrium} with $\propInertia = 4.5e^{-4} Kg m^2$ and $\propDrag = 3.5e^{-7} N m s^2$. The servomotors power consumption is two orders of magnitude smaller and thus negligible.}
    \label{fig:power_ratio}
\end{figure}

If we look at the propellers' power consumption in~\cref{fig:power_ratio}, the three differential methods perform very similarly as they possess similar speed distributions.
On the other hand, \ac{anosecond} slowly pushes the propeller speed up, unnecessarily increasing the power consumption up to $18\%$ more.
The geometric approach is, as expected, the most energy efficient.

\subsection{Flying through singularities}\label{sec:cartwheel}
\begin{figure}[t]
    \includegraphics[width=\columnwidth]{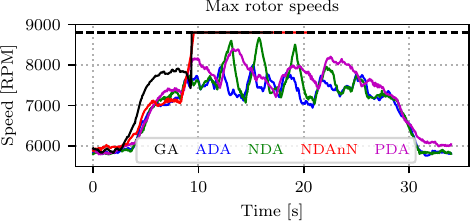}
    \caption{Maximum rotor speeds experienced during the cartwheel trajectory by the different methods. The speed limit allowed by the \acp{ESC} is $8800$ RPM (dashed line). \ac{ageom} and \ac{anosecond} both hit the saturation and diverge, while the other methods keep the propellers away from the limit.}
    \label{fig:cartwheel_rotors}
\end{figure}
As mentioned in~\Cref{sec:geometric_allocation}, geometric allocation methods suffer from singularities which become particularly apparent when the robot is close to a \qty{90}{\degree} pitch or roll angle~\cite{bodie2020singularities}.
Here we show that the proposed differential allocation methods are not affected by these singularities and can still track the desired trajectory even when the robot is at very high angles, unlocking full omnidirectionality.
To further complicate things, in these configurations, the weight of the robot is often compensated by only few propellers, pushing them close to their speed limits.

Consider the cartwheel trajectory in~\cref{fig:vel_time_traj}, where the robot starts in horizontal hover, then pitches up to \qty{90}{\degree} and starts rotating around its body $z$ axis, resembling a rotating cartwheel in the air.
\Ac{ageom} diverges early as it is not able to handle the singularity.
\ac{anosecond} also quickly saturates the propeller speeds and diverges due to its lack of nullspace goals and limit curves.
The other differential methods (\ac{adiff}, \ac{asecond}, \ac{apower}) all successfully complete the trajectory.

All differential allocations try to keep the propellers' speeds low, \ac{adiff} and \ac{asecond} using a secondary objective to do so and \ac{apower} using propeller limit curves, see~\cref{fig:cartwheel_rotors}.

\subsection{Stopping propellers for manipulation}\label{sec:stopping_propellers}
\begin{figure*}[t]
    \includegraphics[width=\textwidth]{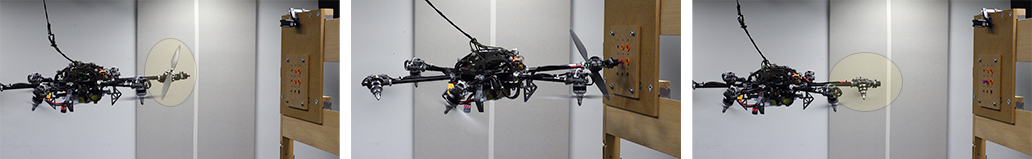}
    \caption{From left to right: the Aerial Robot is hovering in place with the interaction propeller turned off. Then the robot approaches the board and starts rotating the arm to screw the bolt in. Finally, the propeller is re-enabled and the arm aligns with gravity again to balance the actuators' usage. The stopped propeller is highlighted in yellow.}
    \label{fig:stopping_props_pics}
\end{figure*}
\begin{figure}[t]
    \includegraphics[width=\columnwidth]{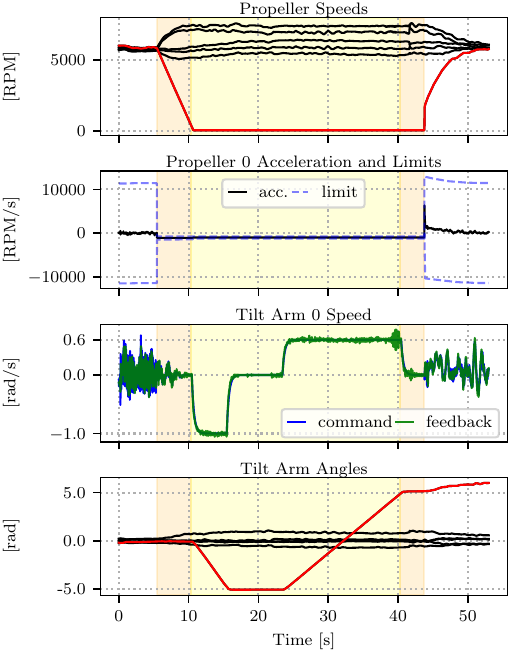}
    \caption{Top to bottom: the speeds of the six propellers (in red the one turned off), the propeller acceleration (and its limits as dashed lines), the speed of the interaction arm, and last the angles of the six arms (in red the manipulation arm). When the acceleration limits are both negative, the propeller slows down until it stops. In the orange shading, the propeller is turned off without explicitly commanding the relative tilt angle. In the yellow shading, the tilt angle is commanded a speed between -1 and \qty{0.6}{\radian\per\second} through the nullspace. In the end, the arm is released and the allocation naturally brings back the propeller to hovering speed to balance the propellers' usage.}
    \label{fig:arm_tilting}
\end{figure}
The possibility of turning off propellers at will opens the door to: \textit{(i)} a safer interaction with the environment, since the propellers close to a surface can be just disabled; \textit{(ii)} additional manipulation capabilities, as the rotating arms can be used for active physical interaction.

Here, we qualitatively show how to use \ac{apower} and the propeller's limit curves to stop one propeller in-flight, summarized in~\cref{fig:stopping_props_pics}.
Meanwhile, we use the nullspace of the allocation to rotate the corresponding arm and screw a bolt into a wall.

In~\cref{fig:arm_tilting}, we present the actuator states and rate commands over time during the manipulation experiment, highlighting the moment when the propeller stops and the arm rotates while approaching the wall.

Note the change in desired propeller acceleration, which is used to stop the propeller, as well as the controlled angular velocity of the arm--both during free flight and throughout the screwing interaction.

We enable this behavior by using the limit curves in~\cref{fig:proposed_limit_curves} to temporarily impose a negative maximum acceleration on the propeller, forcing the allocation to decelerate it until it comes to a stop.
At the end of the interaction, we lift these limits by reapplying the original curves, causing the allocation to naturally restore the balance between propeller speeds.

To track the desired arm speed $\tiltSpeed^*$ in case of unknown friction in the robot's mechanical assembly or with the environment, we use an integral control action in the secondary allocation command
\begin{equation}
    \jointSpeed^* = \tiltSpeed^* + k_I \int \left( \tiltSpeed^* - \tiltSpeed \right)dt,
\end{equation}
where $k_I \in \nR{}$ is the integrator gain and $\tiltSpeed$ the arm rotation velocity feedback from the servomotors.

\section{Conclusion}
In this work, we extend the state-of-the-art differential allocation to include actuator power dynamics and limits.

Experimental results across diverse maneuvers, including a screwing manipulation task, demonstrate improved tracking of high-dynamic trajectories and seamless stopping and restarting of propellers, while keeping the nullspace free to exploit the over-actuation of the system. Additionally, we provide statistical proof that our nullspace-free formulation does not have any performance drawbacks over classical nullspace-based methods.

We introduce propeller limit curves as a compact representation of actuator acceleration limits, enabling a unified treatment of actuator dynamics and user-defined preferences.

The method leverages actuator feedback but avoids reliance on acceleration feedback, which is often unreliable on aerial robots where propeller noise on the IMU and system dynamics are much closer in frequency than on e.g. racing drones. However, our method requires an accurate system model for wrench generation.

Current limitations include the absence of inter-rotor airflow modeling and the assumption of first-order actuator dynamics. Extending the framework to capture these effects is a promising direction for future work.

The proposed formulation is fully compatible with existing wrench-based control architectures and can be integrated without modifying the high-level control stack. It provides a practical and scalable improvement over geometric and differential allocation methods, advancing the capabilities of aerial robotic systems.

\appendix[Computing the propeller limit curves]\label{sec:power_curves_math}
The $9$ coefficients of the power curves can be computed by solving the following linear system of equations

\begin{subequations}\label{eq:power_curves_system}
    \begin{align}
        \max{\propAcc}(\min{\propSpeed}) = c_{0,0}\min{\propSpeed} + c_{0,1}\min{\propSpeed}^2 + c_{0,2}, \\
        c_{0,0}\propSpeedHigh + c_{0,1}\propSpeedHigh^2 + c_{0,2} = c_{1,0}\propSpeedHigh^2 + c_{1,1},    \\
        0 = c_{1,0}\max{\propSpeed}^2 + c_{1,1},                                                          \\
        \propAcc_1 = c_{1,0}\propSpeedHigh^2 + c_{1,1},                                                   \\
        0 = c_{2,0}\min{\propSpeed}^2 + c_{2,1},                                                          \\
        c_{2,0}\propSpeedLow^2 + c_{2,1} = c_{3,0}\propSpeedLow^2 + c_{3,1},                              \\
        \propAcc_2 = c_{2,0}\propSpeedLow^2 + c_{2,1},                                                    \\
        \min{\propAcc}(\max{\propSpeed}) = c_{3,0}\max{\propSpeed}^2 + c_{3,1},                           \\
        0 = c_{0,0}\propSpeed_m + c_{0,1}\propSpeed_m^2 + c_{0,2} + c_{3,0}\propSpeed_m
        \propSpeed_m^2 + c_{3,1}.
    \end{align}
\end{subequations}
This system's solution is unique for the coefficients $c_{i,j}$, given the minimum and maximum propeller speeds $\min{\propSpeed}$, $\max{\propSpeed}$, and accelerations $\max{\propAcc}(\min{\propSpeed})$, $\min{\propAcc}(\max{\propSpeed})$, the speed at which the quadratic drag effect becomes predominant $\propSpeedHigh$ and its acceleration $\max{\propAcc}(\propSpeedHigh)$, the speed where we start saturating the propeller acceleration to avoid reducing its speed below $\propSpeedLow$ and its acceleration $\min{\propAcc}(\propSpeedLow)$, and the desired average propeller speed $\propSpeed_m$.

In our experiments, we have $\propSpeed_m=5800RPM$, while $\min{\propSpeed}=0$ and $\max{\propSpeed}=8700RPM$.
In our experience, the propeller's acceleration starts noticeably decreasing around $\sim8000RPM$.
Thus we use $\propSpeedHigh = 7800RPM$, the $90$\% of the total speed range.
The lower limit $\propSpeedLow$ is a limitation of the ESC, as the speed cannot go below $0$.
For symmetry, we choose $900RPM$, or the $10$\% of the total speed range.
The maximum $\max{\propAcc}(\propSpeedHigh)$ and minimum $\min{\propAcc}(\propSpeedLow)$ accelerations in those points are $80$\% of the absolute maximum and minimum acceleration limits $\max{\propAcc}(\min{\propSpeed})$, $\min{\propAcc}(\max{\propSpeed})$.
Based on our ESC's limitations, we have $\max{\propAcc}(\min{\propSpeed}) = 12000RPM/s$ and $\min{\propAcc}(\max{\propSpeed}) = -14000RPM/s$.

We can then compute the coefficients $c_{i,j}$ from~\eqref{eq:power_curves_system}
with a small matrix inversion.
The values of minimum and maximum acceleration can be changed arbitrarily: both could be negative to stop the propellers in-flight, or restored to their original values to turn them on again.

\begin{figure*}[t]
    \includegraphics[width=\textwidth]{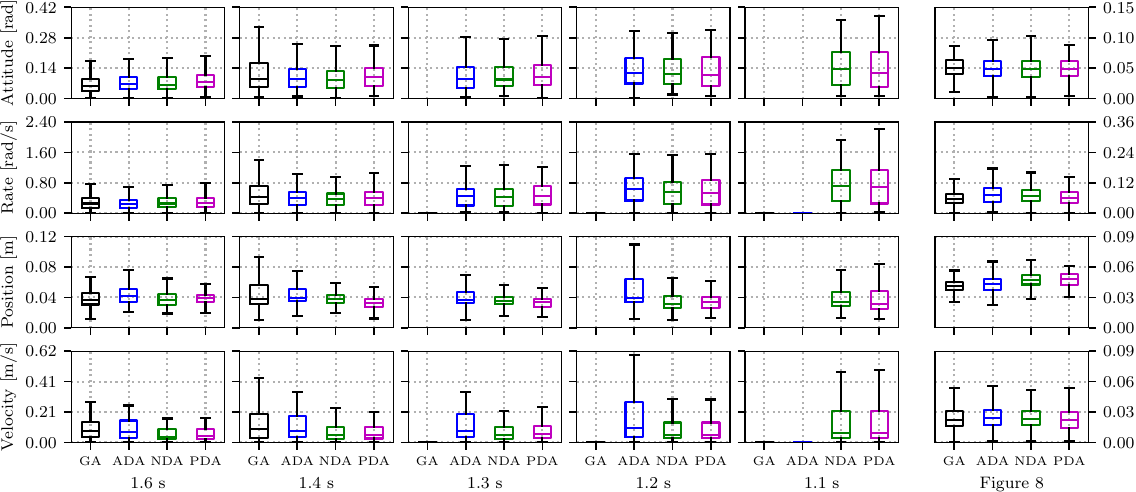}
    \caption{Tracking results from the different allocation methods on several test trajectories. From top to bottom row: Attitude, Angular Velocity, Position and Linear Velocity errors. The five columns on the left show the results with increasingly fast sinusoidal trajectories, from a period of \qty{1.6}{\second} to \qty{1.1}{\second}. The additional column on the right shows the results with a slow Figure-8 trajectory. If no data is present, the corresponding allocation method diverged while following the trajectory.}
    \label{fig:boxplots_errors}
\end{figure*}

\bibliographystyle{IEEEtran}
\bibliography{references.bib}

@article{muellerpairtilt,
  title={PairTilt: Design and Control of an Active Tilt-Rotor Quad-copter for Improved Efficiency and Agility},
  author={Mueller, Jerry Tang Mark W and Tang, Jerry and Mueller, Mark W}
}

@article{sugihara2024beatle,
  title={BEATLE—Self-Reconfigurable Aerial Robot: Design, Control and Experimental Validation},
  author={Sugihara, Junichiro and Zhao, Moju and Nishio, Takuzumi and Okada, Kei and Inaba, Masayuki},
  journal={IEEE/ASME Transactions on Mechatronics},
  year={2024},
  publisher={IEEE}
}

@article{zhao2023versatile,
  title={Versatile articulated aerial robot DRAGON: Aerial manipulation and grasping by vectorable thrust control},
  author={Zhao, Moju and Okada, Kei and Inaba, Masayuki},
  journal={The International Journal of Robotics Research},
  volume={42},
  number={4-5},
  pages={214--248},
  year={2023},
  publisher={SAGE Publications Sage UK: London, England}
}

@article{qin2023design,
  title={Design and flight control of a novel Tilt-Rotor octocopter using passive hinges},
  author={Qin, Zijie and Wei, Jingbo and Cao, Mingzhi and Chen, Baihui and Li, Kaixin and Liu, Kun},
  journal={IEEE Robotics and Automation Letters},
  year={2023},
  publisher={IEEE}
}

@article{nishio2023design,
  title={Design, control, and motion-planning for a root-perching rotor-distributed manipulator},
  author={Nishio, Takuzumi and Zhao, Moju and Okada, Kei and Inaba, Masayuki},
  journal={IEEE Transactions on Robotics},
  year={2023},
  publisher={IEEE}
}

@book{siciliano2008springer,
  title     = {Springer handbook of robotics},
  author    = {Siciliano, Bruno and Khatib, Oussama and Kr{\"o}ger, Torsten},
  volume    = {200},
  year      = {2008},
  publisher = {Springer}
}

@article{Ames2021,
  title     = {Integral control barrier functions for dynamically defined control laws},
  author    = {Ames, Aaron D and Notomista, Gennaro and Wardi, Yorai and Egerstedt, Magnus},
  journal   = {IEEE Control Systems Letters},
  volume    = {5},
  number    = {3},
  pages     = {887--892},
  year      = {2020},
  publisher = {IEEE}
}

@article{cuniato2022power,
  title     = {Power-based safety layer for aerial vehicles in physical interaction using lyapunov exponents},
  author    = {Cuniato, Eugenio and Lawrance, Nicholas and Tognon, Marco and Siegwart, Roland},
  journal   = {IEEE Robotics and Automation Letters},
  volume    = {7},
  number    = {3},
  pages     = {6774--6781},
  year      = {2022},
  publisher = {IEEE}
}

@inproceedings{notomista2020set,
  title        = {A set-theoretic approach to multi-task execution and prioritization},
  author       = {Notomista, Gennaro and Mayya, Siddharth and Selvaggio, Mario and Santos, Mar{\'\i}a and Secchi, Cristian},
  booktitle    = {2020 IEEE International Conference on Robotics and Automation (ICRA)},
  pages        = {9873--9879},
  year         = {2020},
  organization = {IEEE}
}

@article{tognonTRO,
  author  = {Ollero, Anibal and Tognon, Marco and Suarez, Alejandro and Lee, Dongjun and Franchi, Antonio},
  journal = {IEEE Transactions on Robotics},
  title   = {Past, Present, and Future of Aerial Robotic Manipulators},
  year    = {2022},
  volume  = {38},
  number  = {1},
  pages   = {626-645},
  doi     = {10.1109/TRO.2021.3084395}
}

@article{brunner2022mppi,
  author  = {Brunner, Maximilian and Rizzi, Giuseppe and Studiger, Matthias and Siegwart, Roland and Tognon, Marco},
  journal = {IEEE Robotics and Automation Letters},
  title   = {A Planning-and-Control Framework for Aerial Manipulation of Articulated Objects},
  year    = {2022},
  volume  = {7},
  number  = {4},
  pages   = {10689-10696},
  doi     = {10.1109/LRA.2022.3191178}
}

@article{bodie2021tro,
  author  = {Bodie, Karen and Brunner, Maximilian and Pantic, Michael and Walser, Stefan and Pfändler, Patrick and Angst, Ueli and Siegwart, Roland and Nieto, Juan},
  journal = {IEEE Transactions on Robotics},
  title   = {Active Interaction Force Control for Contact-Based Inspection With a Fully Actuated Aerial Vehicle},
  year    = {2021},
  volume  = {37},
  number  = {3},
  pages   = {709-722},
  doi     = {10.1109/TRO.2020.3036623}
}

@inproceedings{brescianini2016omav,
  author    = {Brescianini, Dario and D'Andrea, Raffaello},
  booktitle = {2016 IEEE International Conference on Robotics and Automation (ICRA)},
  title     = {Design, modeling and control of an omni-directional aerial vehicle},
  year      = {2016},
  volume    = {},
  number    = {},
  pges      = {3261-3266},
  doi       = {10.1109/ICRA.2016.7487497}
}

@article{tognon2019tilthex,
  author  = {Tognon, Marco and Chávez, Hermes A. Tello and Gasparin, Enrico and Sablé, Quentin and Bicego, Davide and Mallet, Anthony and Lany, Marc and Santi, Gilles and Revaz, Bernard and Cortés, Juan and Franchi, Antonio},
  journal = {IEEE Robotics and Automation Letters},
  title   = {A Truly-Redundant Aerial Manipulator System With Application to Push-and-Slide Inspection in Industrial Plants},
  year    = {2019},
  volume  = {4},
  number  = {2},
  pages   = {1846-1851},
  doi     = {10.1109/LRA.2019.2895880}
}

@article{park2018odar,
  author  = {Park, Sangyul and Lee, Jeongseob and Ahn, Joonmo and Kim, Myungsin and Her, Jongbeom and Yang, Gi-Hun and Lee, Dongjun},
  journal = {IEEE/ASME Transactions on Mechatronics},
  title   = {ODAR: Aerial Manipulation Platform Enabling Omnidirectional Wrench Generation},
  year    = {2018},
  volume  = {23},
  number  = {4},
  pages   = {1907-1918},
  doi     = {10.1109/TMECH.2018.2848255}
}

@article{ryll2019tilthex,
  author  = {Markus Ryll and Giuseppe Muscio and Francesco Pierri and Elisabetta Cataldi and Gianluca Antonelli and Fabrizio Caccavale and Davide Bicego and Antonio Franchi},
  title   = {6D interaction control with aerial robots: The flying end-effector paradigm},
  journal = {The International Journal of Robotics Research},
  volume  = {38},
  number  = {9},
  pages   = {1045-1062},
  year    = {2019},
  doi     = {10.1177/0278364919856694},
  eprint  = {https://doi.org/10.1177/0278364919856694  }
}

@article{allenspach2020design,
  title     = {Design and optimal control of a tiltrotor micro-aerial vehicle for efficient omnidirectional flight},
  author    = {Allenspach, Mike and Bodie, Karen and Brunner, Maximilian and Rinsoz, Luca and Taylor, Zachary and Kamel, Mina and Siegwart, Roland and Nieto, Juan},
  journal   = {International Journal of Robotics Research},
  volume    = {39},
  number    = {10-11},
  pages     = {1305--1325},
  year      = {2020},
  publisher = {SAGE Publications Sage UK: London, England}
}

@incollection{bodie2020singularities,
  doi       = {10.1007/978-3-030-33950-0_8},
  year      = {2020},
  pages     = {85--95},
  author    = {Karen Bodie and Zachary Taylor and Mina Kamel and Roland Siegwart},
  title     = {Towards Efficient Full Pose Omnidirectionality with Overactuated {MAVs}},
  booktitle = {Springer Proceedings in Advanced Robotics}
}

@inproceedings{cuniato2023minivoliro,
  author    = {Cuniato, Eugenio and Geckeler, Christian and Brunner, Maximilian and Strübin, Dario and Bähler, Elia and Ospelt, Fabian and Tognon, Marco and Mintchev, Stefano and Siegwart, Roland},
  booktitle = {2023 IEEE International Conference on Robotics and Automation (ICRA)},
  title     = {Design and Control of a Micro Overactuated Aerial Robot with an Origami Delta Manipulator},
  year      = {2023},
  volume    = {},
  number    = {},
  pages     = {5352-5358},
  doi       = {10.1109/ICRA48891.2023.10161060}
}

@article{ryll2015fasthex,
  author  = {Ryll, Markus and Bülthoff, Heinrich H. and Giordano, Paolo Robuffo},
  journal = {IEEE Transactions on Control Systems Technology},
  title   = {A Novel Overactuated Quadrotor Unmanned Aerial Vehicle: Modeling, Control, and Experimental Validation},
  year    = {2015},
  volume  = {23},
  number  = {2},
  pages   = {540-556},
  doi     = {10.1109/TCST.2014.2330999}
}

@inproceedings{rajappa2015fasthex,
  author    = {Rajappa, Sujit and Ryll, Markus and Bülthoff, Heinrich H. and Franchi, Antonio},
  booktitle = {2015 IEEE International Conference on Robotics and Automation (ICRA)},
  title     = {Modeling, control and design optimization for a fully-actuated hexarotor aerial vehicle with tilted propellers},
  year      = {2015},
  volume    = {},
  number    = {},
  pages     = {4006-4013},
  doi       = {10.1109/ICRA.2015.7139759}
}

@inproceedings{ryll2016fasthex,
  author    = {Ryll, Markus and Bicego, Davide and Franchi, Antonio},
  booktitle = {2016 IEEE/RSJ International Conference on Intelligent Robots and Systems (IROS)},
  title     = {Modeling and control of FAST-Hex: A fully-actuated by synchronized-tilting hexarotor},
  year      = {2016},
  volume    = {},
  number    = {},
  pages     = {1689-1694},
  doi       = {10.1109/IROS.2016.7759271}
}

@inproceedings{morbidi2018fasthex,
  author    = {Morbidi, Fabio and Bicego, Davide and Ryll, Markus and Franchi, Antonio},
  booktitle = {2018 IEEE/RSJ International Conference on Intelligent Robots and Systems (IROS)},
  title     = {Energy-Efficient Trajectory Generation for a Hexarotor with Dual- Tilting Propellers},
  year      = {2018},
  volume    = {},
  number    = {},
  pages     = {6226-6232},
  doi       = {10.1109/IROS.2018.8594419}
}

@article{ryll2018fasthex,
  title    = {A Truly Redundant Aerial Manipulator exploiting a Multi-directional Thrust Base},
  journal  = {IFAC-PapersOnLine},
  volume   = {51},
  number   = {22},
  pages    = {138-143},
  year     = {2018},
  note     = {12th IFAC Symposium on Robot Control SYROCO 2018},
  issn     = {2405-8963},
  doi      = {https://doi.org/10.1016/j.ifacol.2018.11.531},
  author   = {Markus Ryll and Davide Bicego and Antonio Franchi}
}

@article{ryll2020fasthex,
  author     = {Markus Ryll and
                Davide Bicego and
                Mattia Giurato and
                Marco Lovera and
                Antonio Franchi},
  title      = {FAST-Hex - {A} Morphing Hexarotor: Design, Mechanical Implementation,
                Control and Experimental Validation},
  journal    = {CoRR},
  volume     = {abs/2004.06612},
  year       = {2020},
  eprinttype = {arXiv},
  eprint     = {2004.06612},
  bibsource  = {dblp computer science bibliography, https://dblp.org}
}

@article{falconi2012,
  title    = {Dynamic Model and Control of an Over-actuated Quadrotor UAV},
  journal  = {IFAC Proceedings Volumes},
  volume   = {45},
  number   = {22},
  pages    = {192-197},
  year     = {2012},
  note     = {10th IFAC Symposium on Robot Control},
  issn     = {1474-6670},
  doi      = {https://doi.org/10.3182/20120905-3-HR-2030.00031},
  author   = {Riccardo Falconi and Claudio Melchiorri}
}

@article{johanssen2013allocation,
  title    = {Control allocation—A survey},
  journal  = {Automatica},
  volume   = {49},
  number   = {5},
  pages    = {1087-1103},
  year     = {2013},
  issn     = {0005-1098},
  doi      = {https://doi.org/10.1016/j.automatica.2013.01.035},
  author   = {Tor A. Johansen and Thor I. Fossen}
}

@article{lanegger2023chasing,
  title     = {Chasing millimeters: design, navigation and state estimation for precise in-flight marking on ceilings},
  author    = {Lanegger, Christian and Pantic, Michael and B{\"a}hnemann, Rik and Siegwart, Roland and Ott, Lionel},
  journal   = {Autonomous Robots},
  volume    = {47},
  number    = {8},
  pages     = {1405--1418},
  year      = {2023},
  publisher = {Springer}
}

@article{fabrizio2015nullspace,
  author={Flacco, Fabrizio and De Luca, Alessandro and Khatib, Oussama},
  journal={IEEE Transactions on Robotics}, 
  title={Control of Redundant Robots Under Hard Joint Constraints: Saturation in the Null Space}, 
  year={2015},
  volume={31},
  number={3},
  pages={637-654},
  doi={10.1109/TRO.2015.2418582}}

@article{sun2022indi,
  author={Sun, Sihao and Romero, Angel and Foehn, Philipp and Kaufmann, Elia and Scaramuzza, Davide},
  journal={IEEE Transactions on Robotics}, 
  title={A Comparative Study of Nonlinear MPC and Differential-Flatness-Based Control for Quadrotor Agile Flight}, 
  year={2022},
  volume={38},
  number={6},
  pages={3357-3373},
  doi={10.1109/TRO.2022.3177279}}

\section{Biography Section}
\vspace{-33pt}
\begin{IEEEbiography}[{\includegraphics[width=1in,height=1.25
                    in,clip,keepaspectratio]{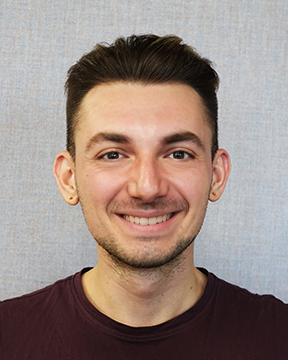}}]{Eugenio Cuniato} is currently a Postdoc at the Autonomous Systems Lab, ETH Zurich, Switzerland. His research focus is on control of aerial manipulators for high-performance aerial physical interaction.
\end{IEEEbiography}

\vspace{-25pt}
\begin{IEEEbiography}
    [{\includegraphics[width=1in,height=1.25in,trim={0 5cm 0 0},clip]{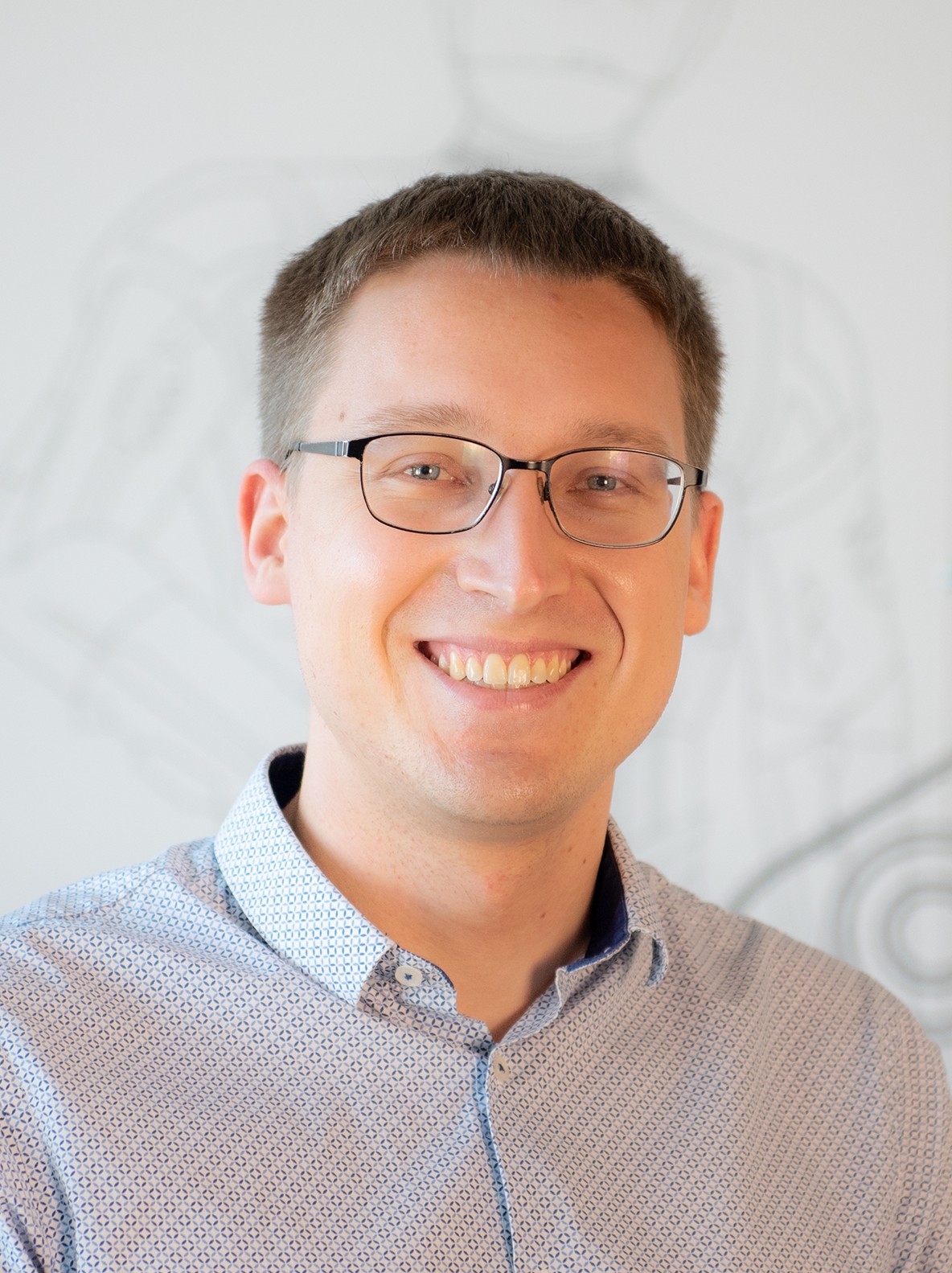}}]{Mike Allenspach} received his Master’s degree in Robotics, Systems, and Control (2020) and his Ph.D. in Robotics (2025) from ETH Zurich, Switzerland. His research focuses on planning and control for aerial manipulation.
\end{IEEEbiography}

\vspace{-25pt}
\begin{IEEEbiography}[{\includegraphics[width=1in,height=1.25in,clip,keepaspectratio]{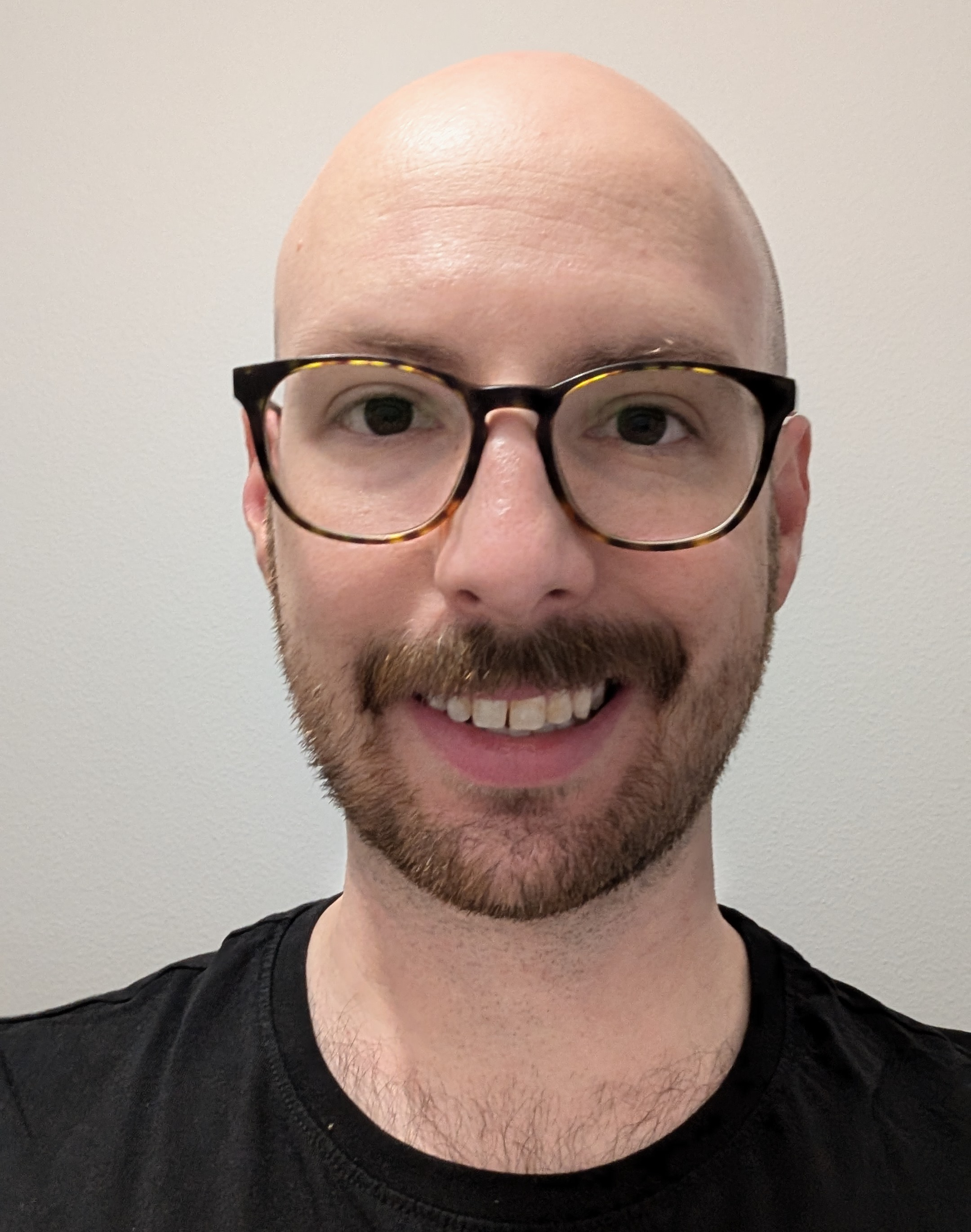}}]{Thomas Stastny} is currently a Senior Researcher in the Autonomous Systems Lab (ASL) at ETH Zurich focusing on learning-based perception and control for omnidirectional aerial manipulators.
\end{IEEEbiography}

\vspace{-25pt}
\begin{IEEEbiography}[{\includegraphics[width=1in,height=1.25in,clip,keepaspectratio]{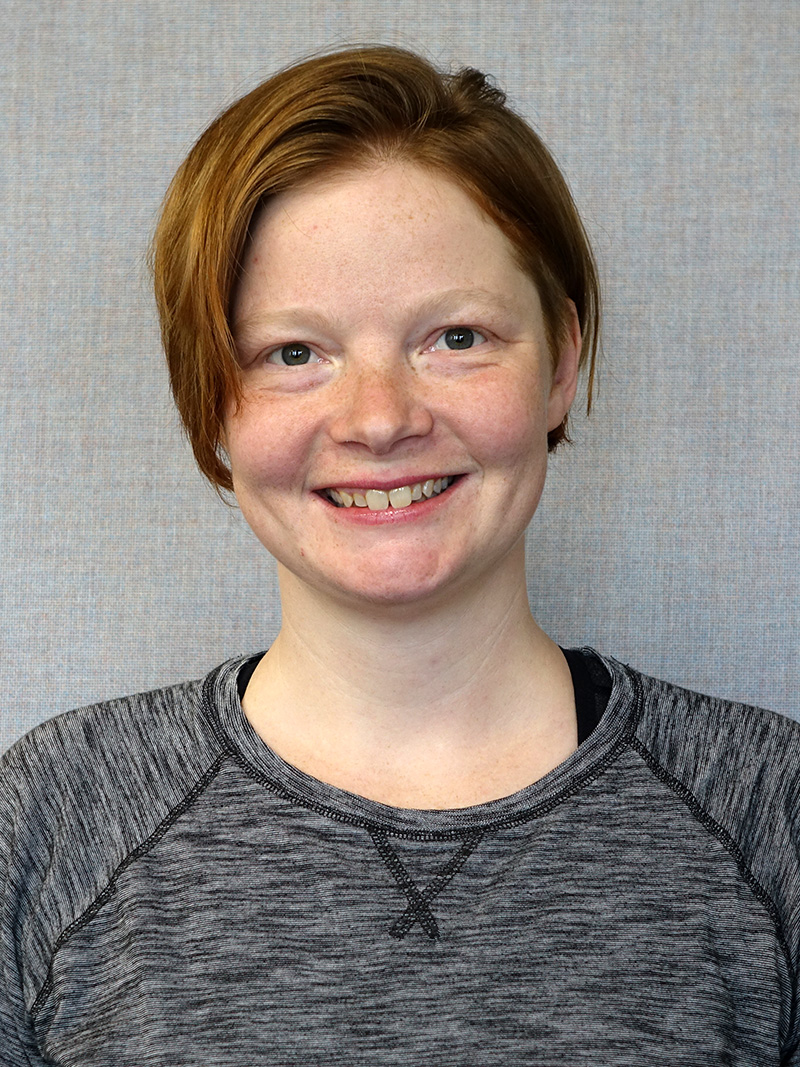}}]{Helen Oleynikova} is also a Senior Researcher in the Autonomous Systems Lab (ASL) at ETH Zurich focusing on volumetric mapping and planning for both manipulation and flying robots.
\end{IEEEbiography}

\begin{IEEEbiography}[{\includegraphics[width=1in,height=1.25in,clip,keepaspectratio]{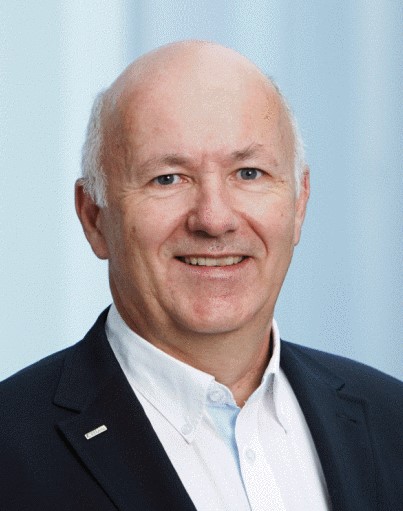}}]{Roland Siegwart}
    is professor for autonomous mobile robots at ETH Zurich, founding co-director of the technology transfer center Wyss Zurich and board member of multiple high tech companies. He was professor at EPFL Lausanne (1996 – 2006), held visiting positions at Stanford University and NASA Ames and was Vice President of ETH Zurich (2010-2014). His interests are in the design, control and navigation of flying, wheeled and walking robots.

\end{IEEEbiography}

\begin{IEEEbiography}[{\includegraphics[width=1in,height=1.25in,clip,keepaspectratio]{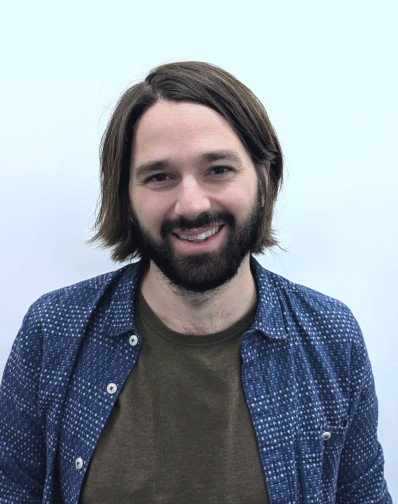}}]{Michael Pantic}
    is a Senior Researcher at the Autonomous Systems Lab (ASL) at ETH Zurich, where he is leading the research on aerial robots. In his own research, he focuses on system architecture, perception and navigation for aerial robots.

\end{IEEEbiography}

\vfill

\end{document}